\documentclass[a4paper,fleqn]{cas-dc}

\usepackage{float}
\usepackage{verbatim} %comments
\usepackage{tabularx}
\usepackage{caption}
\usepackage{newfloat}
\usepackage{threeparttable}
\usepackage[numbers]{natbib}
\usepackage{hyperref}

\DeclareFloatingEnvironment[name={Supplementary Figure}]{suppfigure}
\usepackage{amsmath,amssymb}
\usepackage{nccmath}

\usepackage{multirow}
\usepackage{lineno}

\usepackage{setspace}
\usepackage{subcaption}
\hypersetup{
  colorlinks=true,
  linkcolor=blue,   % 设置链接颜色为蓝色
  urlcolor=red      % 设置URL颜色为红色
}

\restylefloat{figure}
\def\tsc#1{\csdef{#1}{\textsc{\lowercase{#1}}\xspace}}
\tsc{WGM}
\tsc{QE}
\tsc{EP}
\tsc{PMS}
\tsc{BEC}
\tsc{DE}
\begin{document}
\let\WriteBookmarks\relax
\def\floatpagepagefraction{1}
\def\textpagefraction{.001}

% Short title
\shorttitle{Computer Methods and Programs in Biomedicine}

% Short author
\shortauthors{Junming REN et~al.}

% Main title of the paper
\title [mode = title]{EPIDetect: Video-based convulsive seizure detection in
chronic epilepsy mouse model for anti-epilepsy drug screening}                      
% Title footnote mark
% eg: \tnotemark[1]
% \tnotemark[1,2]

% Title footnote 1.
% eg: \tnotetext[1]{Title footnote text}
% \tnotetext[<tnote number>]{<tnote text>} 
% \tnotetext[1]{This document is the results of the research
%    project funded by the National Science Foundation.}

% \tnotetext[2]{The second title footnote which is a longer text matter
%    to fill through the whole text width and overflow into
%    another line in the footnotes area of the first page.}

% First author
%
% Options: Use if required
% eg: \author[1,3]{Author Name}[type=editor,
%       style=chinese,
%       auid=000,
%       bioid=1,
%       prefix=Sir,
%       orcid=0000-0000-0000-0000,
%       facebook=<facebook id>,
%       twitter=<twitter id>,
%       linkedin=<linkedin id>,
%       gplus=<gplus id>]
\author[1]{Junming REN}[
% type=editor,
                        auid=000,bioid=1,
                        % prefix=,
                        % role=PhD Student,
                        orcid=0000-0003-1693-7845]

% Corresponding author indication

% Footnote of the first author
\fnmark[1]

% Email id of the first author
\ead{junminren2-c@my.cityu.edu.hk}

% URL of the first author
% \ead[url]{www.cvr.cc, cvr@sayahna.org}

%  Credit authorship
\credit{Conceptualization of this study, Methodology, Software}

% Second author
\author[1]{Zhoujian XIAO}[style=chinese,
auid=000,bioid=1,
                        % prefix=,
                        % role=PhD Student,
                        orcid=0000-0002-8316-6051]
\fnmark[1]
\ead{zhouxiao-c@my.cityu.edu.hk}
% \ead[URL]{www.sayahna.org}

% Third author
\author[1]{Yujia ZHANG}[style=chinese,
auid=000,bioid=1,
                        % prefix=,
                        % role=PhD Student,
                        orcid=0000-0003-3991-7388]
\fnmark[1]
\ead{yzhang2383-c@my.cityu.edu.hk}
% \ead[URL]{www.sayahna.org}

% Fourth author
\author[1]{Yujie YANG}[style=chinese,
auid=000,bioid=1,
                        % prefix=,
                        % role=PhD Student,
                        orcid=0000-0002-7202-3776]
\ead{yujieyang4-c@my.cityu.edu.hk}
\credit{Data curation, Writing - Original draft preparation}

% Fifth  author
\author[1]{Ling HE}[style=chinese,
auid=000,bioid=1,
                        % prefix=,
                        % role=PhD Student,
                        orcid=0000-0001-8723-7702]
\ead{linghe5@cityu.edu.hk}
\credit{Data curation, Writing - Original draft preparation}

% 6th  author
\author[1]{Lijia CHE}[style=chinese,
auid=000,bioid=1,
                        % prefix=,
                        % role=PhD Student,
                        orcid=0000-0002-8784-2897]
\ead{lijiache2-c@my.cityu.edu.hk}
\credit{Data curation, Writing - Original draft preparation}

% 7th  author
\author[1]{Ezra Yoon}[style=chinese,
auid=000,bioid=1,
                        % prefix=,
                        % role=PhD Student,
                        orcid=0000-0002-8216-680X]
\ead{ezrayoon@cityu.edu.hk}
\credit{Data curation, Writing - Original draft preparation}

% 8th  author
\author[1]{Stephen Temitayo Bello}[style=chinese,
auid=000,bioid=1,
                        % prefix=,
                        % role=PhD Student,
                        orcid=0000-0002-4730-4349]
\ead{stbello2-c@my.cityu.edu.hk}
\credit{Data curation, Writing - Original draft preparation}

% 9th  author
\author[1]{Mengfan ZHANG}[style=chinese,
auid=000,bioid=1,
                        % prefix=,
                        % role=PhD Student,
                        orcid=0000-0002-5320-5742]
\ead{mzhang447-c@my.cityu.edu.hk}
\credit{Data curation, Writing - Original draft preparation}

% 10th  author
\author[1]{Xi CHEN}[style=chinese,
auid=000,bioid=1,
                        % prefix=,
                        % role=PhD Student,
                        orcid=0000-0002-2144-6584]
\ead{xi.chen@cityu.edu.hk}
\credit{Data curation, Writing - Original draft preparation}

% 11th  author
\author[1]{Dapeng WU}[style=chinese,
auid=000,bioid=1,
                        % prefix=,
                        % role=PhD Student,
                        orcid=0000-0003-1755-0183]
\ead{dapengwu@cityu.edu.hk}
\credit{Data curation, Writing - Original draft preparation}

% 12th  author
\author[2]{Micky Tortorella}[style=chinese,
auid=000,bioid=1,
                        % prefix=,
                        % role=PhD Student,
                        orcid=0000-0002-1010-0174]
\ead{m.tortorella@crmh-cas.org.hk}
\credit{Data curation, Writing - Original draft preparation}

% 13th  author
\author[1]{Jufang HE}[style=chinese,
auid=000,bioid=1,
                        % prefix=,
                        % role=PhD Student,
                        orcid=0000-0002-4288-5957]
\ead{jufanghe@cityu.edu.hk}
\cormark[1]
\credit{Data curation, Writing - Original draft preparation}

% Address/affiliation
\affiliation[1]{organization={City University of Hong Kong},
    addressline={83 Tat Chee Avenue}, 
    city={Kowloon},
    % citysep={}, % Uncomment if no comma needed between city and postcode
    postcode={999077}, 
    state={},
    country={Hong Kong}}
% Address/affiliation
\affiliation[2]{organization={Center for Regenerative Medicine and Health, Hong Kong Institute of Science \&
Innovation, Chinese Academy of Sciences, Hong Kong,},
    addressline={15 Science Park West Avenue}, 
    city={New Territories},
    % citysep={}, % Uncomment if no comma needed between city and postcode
    postcode={999077}, 
    state={Hong Kong}
    % country={Hong Kong}
    }
% \affiliation[3]{organization={STM Document Engineering Pvt Ltd.},
%     addressline={Mepukada}, 
%     city={Malayinkil},
%     % citysep={}, % Uncomment if no comma needed between city and postcode
%     postcode={695571}, 
%     state={Trivandrum},
%     country={India}}

% Corresponding author text
\cortext[cor1]{Corresponding author}
% \cortext[cor2]{Principal corresponding author}

% Footnote text
\fntext[fn1]{Contributed equally.}
% For a title note without a number/mark
% \nonumnote{This note has no numbers. In this work we demonstrate $a_b$
%   the formation Y\_1 of a new type of polariton on the interface
%   between a cuprous oxide slab and a polystyrene micro-sphere placed
%   on the slab.
%   }

% Here goes the abstract
\begin{abstract}
\phantom{}\hspace{-0.3em}
\textit{Background:}  
In the preclinical translational studies, drug candidates with remarkable anti-epileptic efficacy demonstrate long-term suppression of spontaneous recurrent seizures (SRSs), particularly convulsive seizures (CSs), in mouse models of chronic epilepsy. However, the current methods for monitoring CSs have limitations in terms of invasiveness, specific laboratory settings, high cost, and complex operation, which hinder drug screening efforts. In this study, a camera-based system for automated detection of CSs in chronically epileptic mice is first established to screen potential anti-epilepsy drugs. \\ \textit{Methods:}  To achieve this goal, a novel deep learning-based framework is developed called EPIDetect, which integrates both an action recognition network (ARN) and an object detection network (ODN) to train a large-scale trimmed video dataset with chronic epilepsy mice in home cages. EPIDetect equipped with advanced decision-making techniques is evaluated in three different datasets, including validation of trained models, assessment of robustness across different new home-caged mice, and examination of generalizability to chambered mice. Furthermore, EPIDetect is applied in practical experiments to assess new drug efficacy and to compare its performance with that of human experts. \\ \textit{Results:} EPIDetect demonstrates outstanding performance in the test dataset for robustness (recall: 99.2\%, precision: 98.8\%, and F1 score: 99.0\%) and generalization (recall: 91.3\%, precision: 95.5\%, and F1 score: 93.3\%). Remarkably, EPIDetect found a small molecular antagonist to suppress CSs and achieved a 3.7\% improvement in the hint rate of CSs with only 1/175 of the time consumption compared with human experts. \\ \textit{Conclusion:} These experimental findings suggest that the EPIDetect platform has the potential to serve as a high-efficient tool for in vivo anti-epilepsy drug screening, replacing traditional methods.

% \noindent\texttt{\textbackslash begin{abstract}} \dots 
% \texttt{\textbackslash end{abstract}} and
% \verb+\begin{keyword}+ \verb+...+ \verb+\end{keyword}+ 
% which
% contain the abstract and keywords respectively. 

% \noindent Each keyword shall be separated by a \verb+\sep+ command.
\end{abstract}

% Use if graphical abstract is present
% \begin{graphicalabstract}
% \includegraphics{figs/grabs.pdf}
% \end{graphicalabstract}

% Research highlights
% \begin{highlights}
% \item Research highlights item 1
% \item Research highlights item 2
% \item Research highlights item 3
% \end{highlights}

% Keywords
% Each keyword is seperated by \sep
\begin{keywords}
Chronically Epileptic Mice \sep Convulsive Seizures \sep Deep Learning\sep EPIDetect \sep Anti-epilepsy Drug Screening \sep Action Recognition Network
\end{keywords}

\maketitle

\section{Introduction}
Epilepsy is a global health concern, affecting approximately 65 million people worldwide, with an alarming five million new cases diagnosed annually \cite{fiest2017prevalence}. Despite advances in medical science, treatment options for individuals, both children and adults, living with epilepsy remain suboptimal, leaving many patients grappling with uncontrolled seizures, and some facing the dire risk of sudden unexpected death in epilepsy (SUDEP) \cite{tomson2005sudden}. Thus, there is an urgent need to develop more effective anti-epilepsy drugs aimed at mitigating the risk of SUDEP from refractory epilepsy. Spontaneous recurrent seizures (SRSs) are a hallmark feature of refractory epilepsy. Among SRSs, convulsive seizures (CSs) represent one of the most severe manifestations, characterized by overt and observable convulsive behaviors \cite{kang2017equivocal}. Accurate detection and continuous monitoring of CSs can serve as a pivotal indicator of the effectiveness of the drug during long-term therapy. Current preclinical translational research in the field of epilepsy heavily relies on continuous Video-electroencephalogram (video-EEG) monitoring as a diagnostic tool \cite{gschwind2023hidden}. Video-EEG not only captures the seizure burden across various timescales, including ultradian \cite{stirling2021seizure} and circadian dynamics \cite{karoly2021cycles}, but it also allows for the study of disease progression (epileptogenesis) \cite{mazzuferi2012rapid}. However, the limitations of this method lie in its confinement to specialized monitoring units and the invasive nature of the procedures, which restricts the natural behaviors of mice under investigation \cite{yang2021video}. Consequently, the labor-intensive nature of video-EEG monitoring hinders the exploration of long-term changes in seizure behaviors throughout the course of treatment. observation and quantification of CSs from video is a non-invasive and direct method to test drug efficacy. However, it needs at least 147 days for manual observation (three weeks for seven mice per experimental group), which is non-objective and energy-exhaust \cite{beesley2020d}. The development of automated techniques for the detection of CSs in home-caged mice has the potential to enhance monitoring capabilities and significantly reduce the time and labor investments required for the screening and evaluation of long-term video data.
Moreover, there exists a growing desire to identify and monitor seizures using video data exclusively, as this approach eliminates the need for physical contact with the subjects and can be readily implemented, often utilizing pre-existing video hardware in various settings \cite{maekawa2020deep}. 
In this study, we introduce an automated method designed to directly detect convulsive behaviors in lengthy videos recorded within home-caged environments. Our approach is underpinned by a supervised action recognition network (ARN) framework that is complemented by an object detection network (ODN), collectively known as EPIDetect. This innovative system has been specifically developed to facilitate the detection of seizure events, particularly convulsions.  

We highlight the distinct contributions of this research:
\begin{enumerate}
\item This study represents the first attempt to automate the detection of convulsive seizures in chronically epileptic mice models. Combining supervised machine learning and object detection techniques, as well as leveraging transfer learning from large models, effectively addresses the highly dynamic nature of seizure behaviors.

\item The integration of model ensembling and threshold adjustment techniques significantly enhances the performance of seizure detection in unseen subjects and diverse environments. This advancement holds great promise for reliable application in drug efficacy testing by quantifying changes in seizure frequency.

\item This study provides the first large dataset for long-term recording of freely behaving epileptic mice. It enables researchers to analyze additional potential pharmacological features and accelerate drug development. Furthermore, it serves as a benchmark for algorithm development in various drug efficacy tasks.

\end{enumerate}

% All the above packages are part of any
% standard \LaTeX{} installation.
% Therefore, the users need not be
% bothered about downloading any extra packages.

\section{Literature review}

Historically, the automatic detection of specific actions relied heavily on manual, hand-crafted approaches utilizing depth data \cite{jalal2020automatic}, skeletal information \cite{yan2018spatial}, and hybrid features \cite{ramanujam2021human}. However, the performance of these methods has proven to be inconsistent due to the inherent challenges posed by real-world environments \cite{nagabandi2018learning}. With the evolution of data-driven techniques, non-parametric methods, notably Convolutional Neural Networks (CNNs), have been increasingly applied in behavior detection \cite{yao2019review}. Two-dimensional-CNNs (2dCNN) excel at extracting and learning appearance information from various sources within a single frame, such as RGB data \cite{munro2020multi}, optical flow \cite{sevilla2019integration}, pose \cite{xiaohan2015joint}, and trajectory \cite{zhao2018trajectory}. Nonetheless, when dealing with convulsive seizures, which encompass specific patterns like myoclonic seizures \cite{baykan2008myoclonic}, clonic seizures \cite{theodore1994secondarily}, and tonic-clonic seizures \cite{van2019ptz}, it becomes evident that they contain rich spatial and temporal information. Consequently, single-frame-based methods face significant challenges when it comes to accurately detecting seizure behaviors. Compared to traditional 2dCNNs, the integration of CNNs with temporal convolutional networks (TCNs), such as two-stream spatial-temporal networks \cite{feichtenhofer2019slowfast} or three-dimensional CNNs (3dCNN) \cite{jiang2019stm}, has demonstrated the capacity to effectively capture these complex behaviors. This is substantiated by their state-of-the-art performance on large-scale human action datasets like Kinetics \cite{carreira2017quo}. Furthermore, recent research has shown that pre-trained models using transfer learning consistently outperform models trained from scratch \cite{salman2020adversarially}.
While video-based spontaneous behavior analysis in mice has been extensively researched in the past, recent advances have harnessed the power of Deep Neural Networks (DNNs) for the behavioral analysis of laboratory animals \cite{mathis2020deep}. These approaches have achieved remarkable performance by focusing on spatial feature classification through techniques like pose estimation and manual definition. For example, DeepLabcut generates skeletal representations of mice based on pose features, further categorizing their behavior subtypes \cite{mathis2018deeplabcut}. Hierarchical decomposition has been applied to spatial-temporal skeletons for behavior classification \cite{huang2021hierarchical}. However, the field of seizure detection using automated behavioral analysis in mice remains relatively underexplored. For decades, the primary tool for analyzing seizure behaviors in mice has been SeizureScan, a system that classifies seizure behavior through manual feature extraction \cite{boillot2014glutamatergic}. However, it is limited to specific experimental settings and primarily captures behaviors in rats. There has been an instance where object detection was used to locate kainate-induced seizure rat models and a 2dCNN was employed to classify their wet-dog shaking behaviors during spontaneous recurrent seizures using three different cameras (precision: 0.91, recall: 0.86) \cite{negrete2023multi}. Nevertheless, the reliance on frame-based detection has limited its performance, particularly in the context of continuous CSs, and the detection of smaller animals, such as mice, has proven to be more challenging. Recent developments have explored motion sequences \cite{wiltschko2020revealing} using point clouds generated from depth cameras to construct 3D features in epileptic mice, enabling the differentiation of varying degrees of seizure behaviors with an F1 score of 0.6 \cite{voskobiynyk2023ai}. However, this approach is often constrained to experimental chambers and lacks the capability to provide continuous monitoring over a 24-hour period while ensuring animal welfare.
Therefore, we developed a novel approach using 3dCNN-based ARN with object localization to detect convulsive seizures and to quantify the seizure frequency for the application with anti-epilepsy drug screening.

\section{Methods}

\subsection{Technical Background}

\begin{figure*}[t]
  \begin{minipage}{0.4\textwidth}
    \centering
    \begin{subfigure}[t]{\linewidth}
      \includegraphics[height=6cm, width=\linewidth]{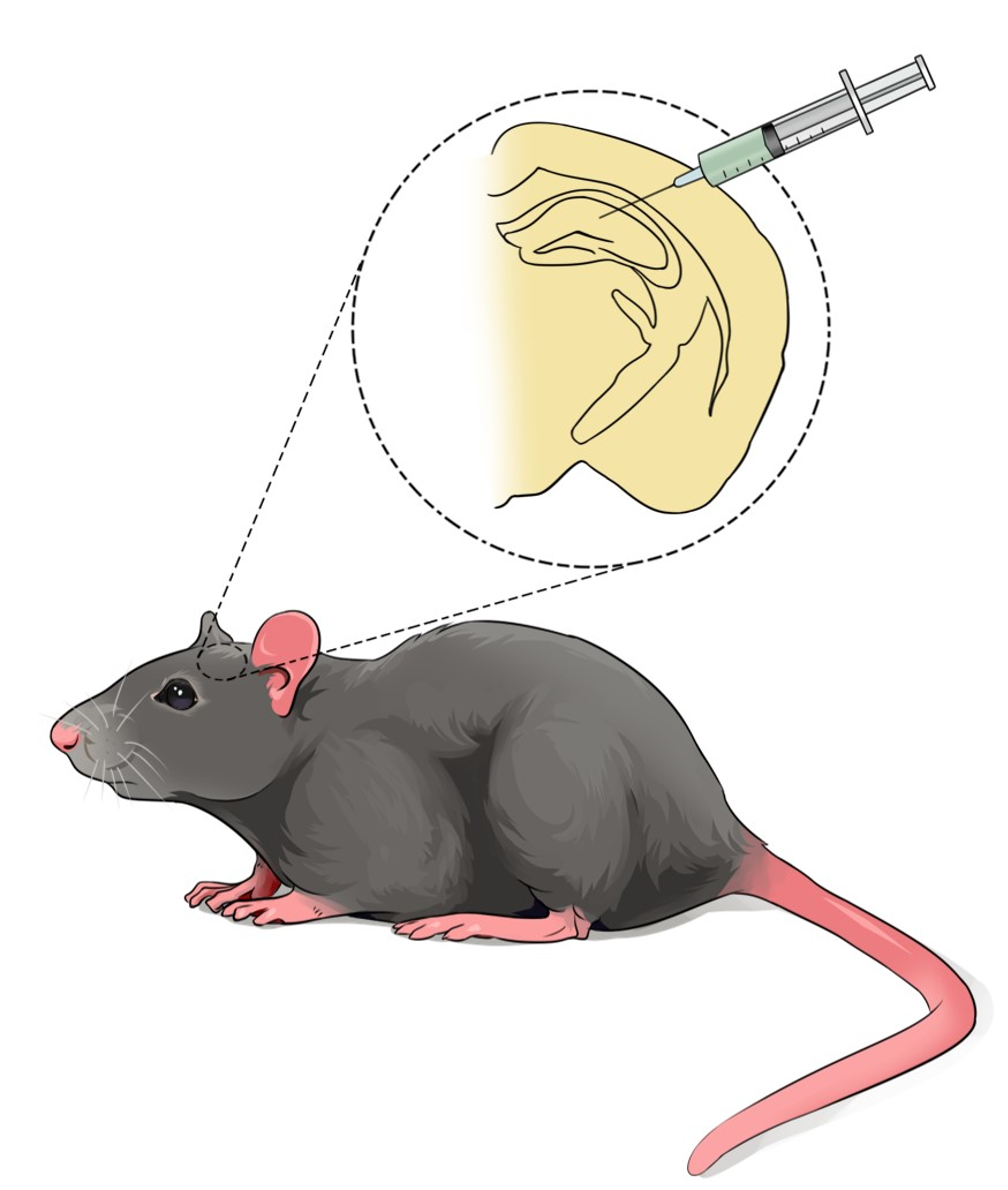}
      \caption{Animal Model}
      \label{fig:animal_model}
    \end{subfigure}
  \end{minipage}
  \hfill
  \begin{minipage}{0.5\textwidth}
    \centering
    \begin{subfigure}[t]{\linewidth}
      \includegraphics[height=6cm, width=\linewidth]{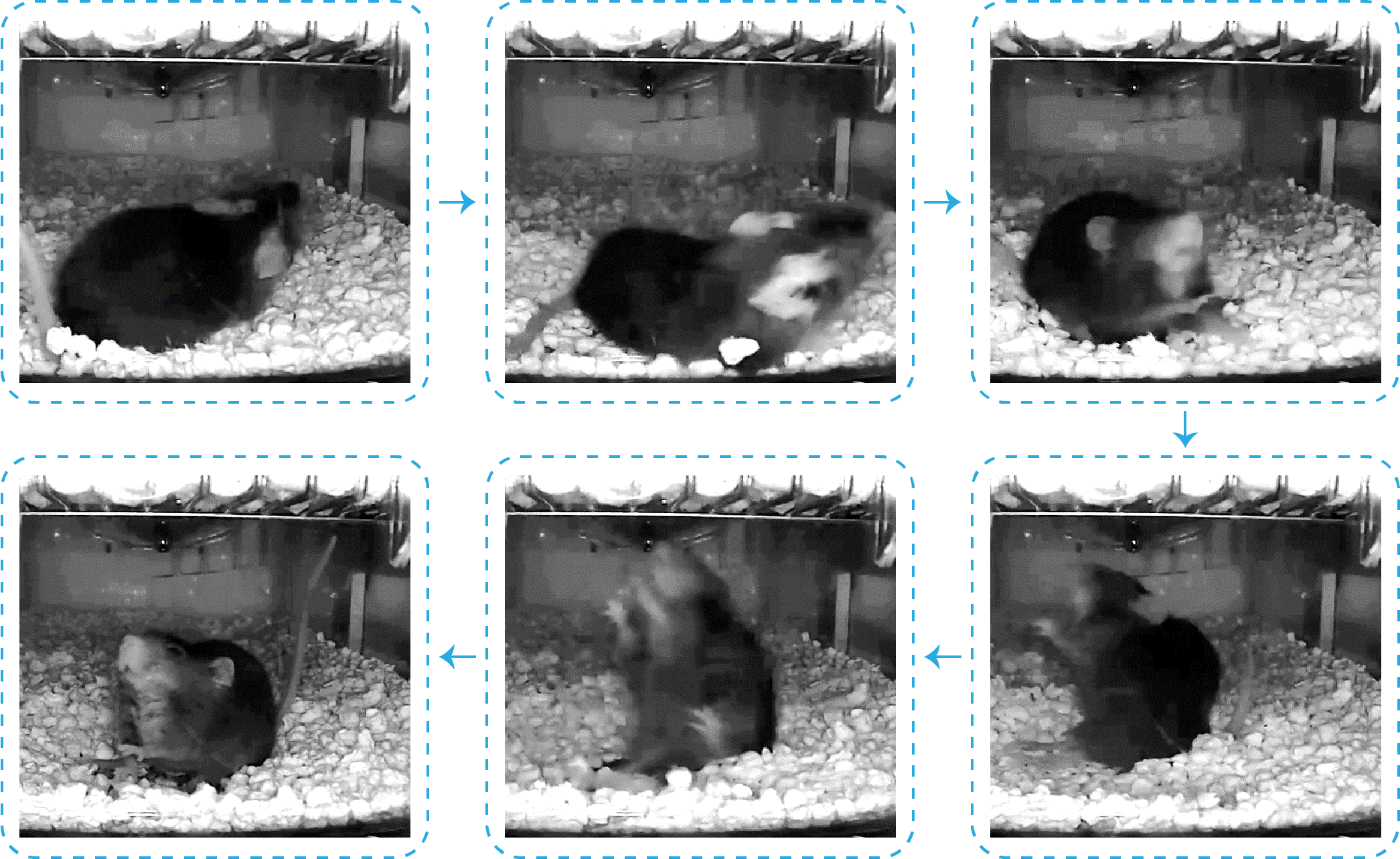}
      \caption{Seizure action}
      \label{fig:seizure_action}
    \end{subfigure}
  \end{minipage}
  \caption{Schematic of unilateral IHKA model of chronic TLE (a) and convulsive behavior characteristic of the spontaneous recurrent seizure (b).}
  \label{fig:injection}
\end{figure*}

For the study of acquired epilepsy, we employed the unilateral
intrahippocampal kainic acid (IHKA) model, a well-established model for
chronic temporal lobe epilepsy (TLE) \cite{levesque2013kainic}. This model is renowned for its
ability to faithfully replicate key histological, electrographic, and
cognitive features that closely mirror human TLE, which stands as the
most prevalent form of intractable epilepsy among adults. The IHKA model entails the precise injection of KA into the hippocampus,
thereby inducing neuronal damage that effectively mimics the
pathological features associated with SRSs (see \textbf{Fig. \ref{fig:injection}}a).

For the intrahippocampal drug injection, a total volume of 650
nanoliters (nl) of KA solution (0.3 mg/ml, Tocris) was delivered using a
Nanoliter Injector (World Precision Instruments, USA). The injection
site was meticulously targeted to the CA1 region of the right dorsal
hippocampus, with coordinates positioned at -2.06 mm posterior to
bregma, -1.80 mm to the midline, and -1.60 mm to the dura. The injection
was administered at a controlled rate of 30 nl/min. The experimental
subjects were C57BL/6 mice, aged between 6 to 10 weeks, and they were
appropriately anaesthetized before receiving the controlled KA solution
via injection. Subsequent to the injection, the mice exhibited early
symptomatic seizures, progressing to spontaneous recurrent seizures.

In the context of spontaneous recurrent seizures, it is pertinent to
note that convulsive seizures represent the most readily observable and
severe form of seizure activity. Convulsive seizures within this model
are characterized by the tonic phase, marked by muscle stiffness,
followed by the clonic phase, characterized by rhythmic jerking
movements (see \textbf{Fig. \ref{fig:injection}}b). These convulsive seizures are
emblematic of severe seizure activity, surpassing the earlier stages of
the Racine Scale \cite{luttjohann2009revised}. The utilization of the Racine Scale serves to
standardize the description of seizure behavior in animal models,
facilitating meaningful comparisons of results across various research
studies.

It is worth emphasizing that all experimental procedures were conducted
in strict accordance with the ethical guidelines and were duly approved
by the Animal Subjects Ethics Sub-Committee of the City University of
Hong Kong.

\subsection{Experiment Environment Set-up}

The mice used in this study were housed in standard ICV cages ($45 \text{cm} *
30 \text{cm} * 25 \text{cm}$) within the Laboratory Animal Research Unit (LARU). The
housing environment was maintained with a 12-hour light and 12-hour dark
cycle, and strict control over climatic conditions, including
temperature, humidity, and sterility, was ensured. Furthermore, the mice
had continuous access to food and water, which was managed by
professional staff at LARU. Their well-being was maintained through free
access to food and water, allowing them to exhibit normal behaviors. All
experimental procedures were conducted in strict compliance with the
regulations set forth by the City University Animal Welfare and Ethical
Review Body.

\begin{figure*}[t]
  \begin{minipage}{0.5\textwidth}
    \centering
    \begin{subfigure}[t]{\linewidth}
    \includegraphics[height=4cm, width=\linewidth]{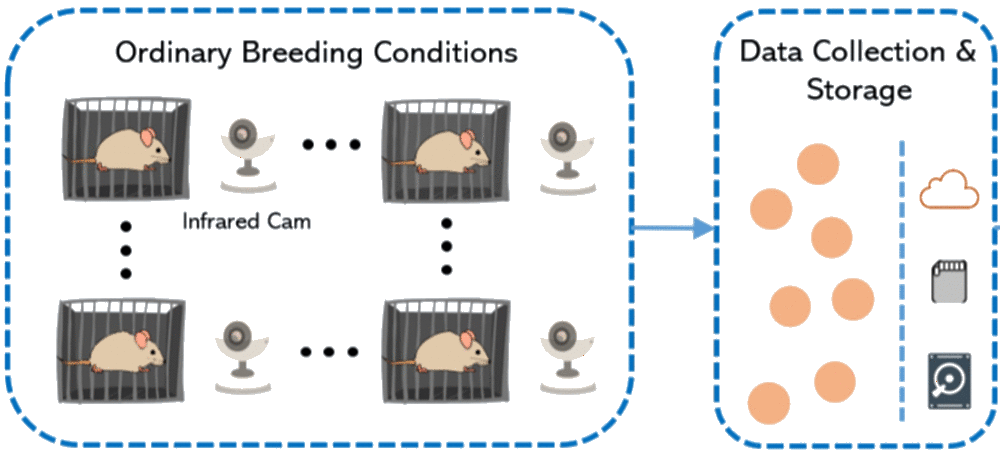}
    \caption{Schematic}
    \end{subfigure}
  \end{minipage}%
  \hspace{1cm} % Add desired horizontal space here
  \begin{minipage}{0.4\textwidth}
    \centering
    \begin{subfigure}[t]{\linewidth}
    \includegraphics[height=4cm, width=\linewidth]{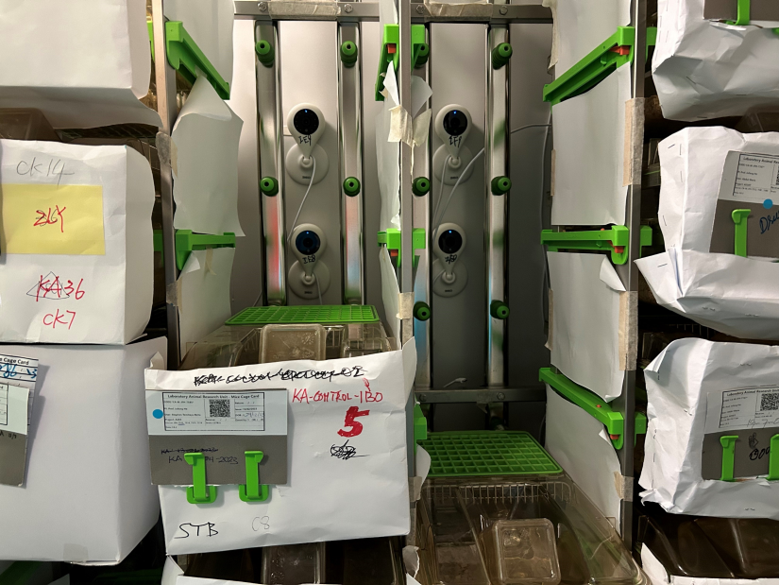}
    \caption{Environment}
    \end{subfigure}
  \end{minipage}
  \caption{Experimental environment setting and data acquisition}
  \label{fig:environment}
\end{figure*}

Data capture was accomplished using a standard single RGB camera with a
horizontal view spanning from the front to the back of the experimental
setup, which spanned 24 hours (see \textbf{Fig. \ref{fig:environment}}). For recordings
within the experimental chambers, each mouse was recorded using a
high-resolution 1080P camera with a single view (from front to back)
while placed within an acrylic test box (size) located inside a
light-and-sound-proof chamber. This recording was conducted for a
duration of two hours per day.

\subsection{Establishment of the Methodology}

\begin{figure*}[t]
    \centering
    \includegraphics[width=\linewidth]{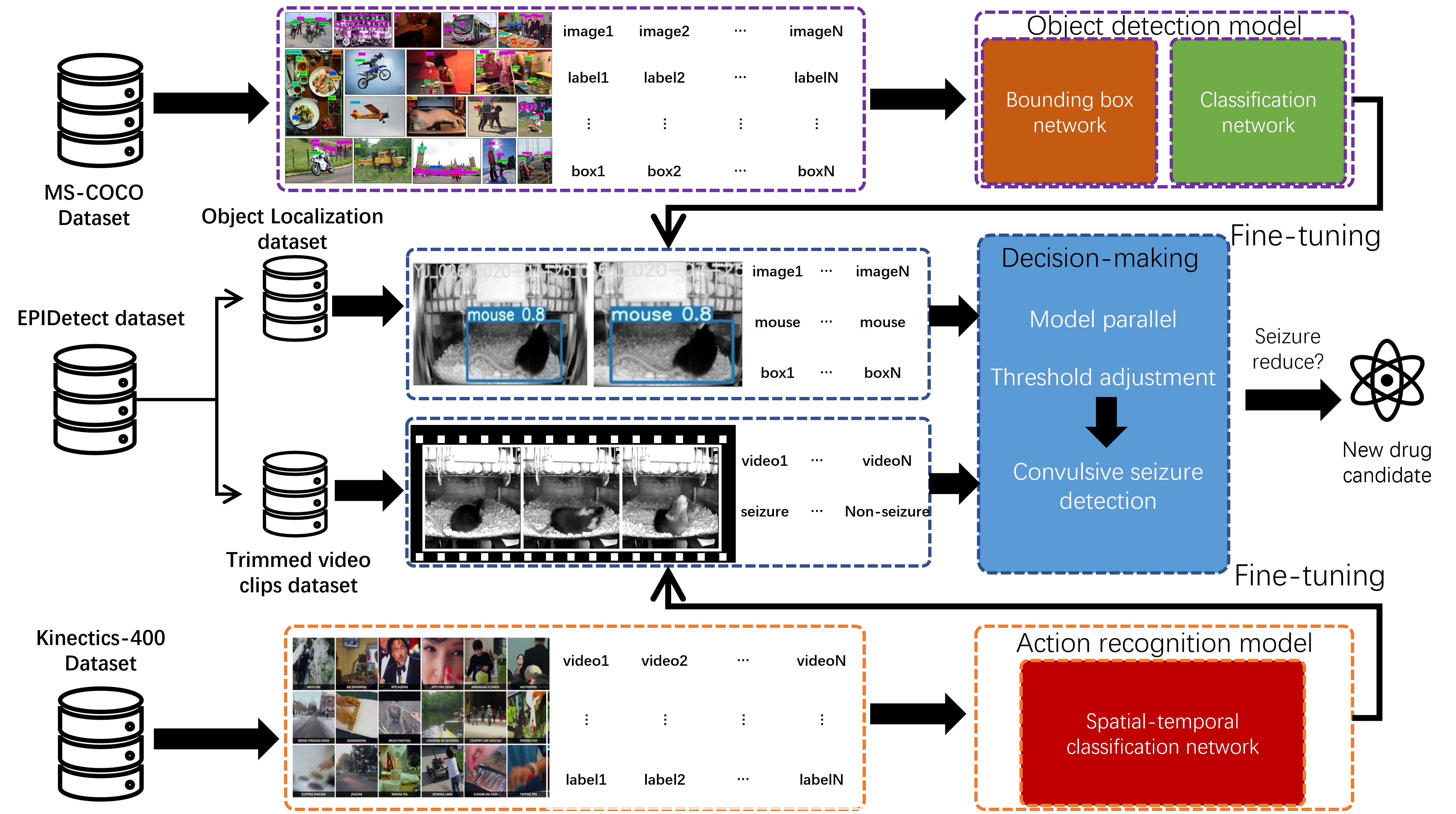}
    \caption{Overview of EPIDetect framework in convulsive behavior detection}
    \label{fig:framework}
\end{figure*}

The EPIDetect fusion architecture, as illustrated in \textbf{Fig. \ref{fig:framework}}, is
designed to facilitate the recognition of epileptic behavior in mice.
This architecture combines two key components: the Object Detection
Network (ODN) and the Action Recognition Network (ARN).

The ODN component is trained and validated using the YOLOv5 framework \cite{redmon2016you}).
It leverages a pre-labeled home-caged dataset, complete with bounding
box annotations. The primary objective of ODN is to accurately identify
and localize epileptic mice within video frames.

In parallel, the ARN component is built upon the three-dimensional ResNext (3d-ResNext) framework, a
state-of-the-art network architecture commonly employed for human action
recognition tasks \cite{crasto2019mars}). In our context, the ARN is tasked with the detection
of epileptic mice within video sequences. When an epileptic mouse is
detected, a square region encompassing the mouse\textquotesingle s
presence is extracted as the input for the subsequent behavior
recognition network.

The next step in the process involves cropping the extracted frames to
facilitate binary classification: distinguishing between seizure and
non-seizure behavior. This classification is performed by the ARN, which
has been fine-tuned for the specific task of seizure detection.

To enhance the performance of the model, advanced
decision-making methods are applied. These include model parallel and threshold adjustment strategies, which collectively refine
the predicted results and improve the overall performance of the system.

% \begin{figure}[H]
%     \centering
%     \includegraphics[width=\linewidth]{Figure3.png}
%     \caption{Overview of EPIDetect framework in convulsive behavior detection}
%     \label{fig:framework}
% \end{figure}

\subsubsection{Object Detection Network}

In our methodology, we employed the YOLOv5, renowned for its expedited meta-network architecture while maintaining a commendable accuracy score. YOLOv5 is structured into four key components: Input, Backbone, Neck, and Prediction. The input component features Mosaic data augmentation, encompassing random zoom, random crop, and random arrangement for stitching, along with adaptive anchor frame calculation to aid in network output prediction frame generation. YOLOv5 introduces adaptive picture zoom to significantly enhance inference speed while minimizing zoom and eliminating black edges. The Backbone component, which includes Focus for sliding operations, utilizes the CSP (Cross Stage Partial Network) structure to address the challenge of large computational costs during inference. The Neck incorporates the
Feature Pyramid Networks and Path Aggregation Networks
modules, facilitating the transfer of information and interaction
between clusters of strong semantic and positional features. For output,
we employ the \(\mathcal{L}_{CIoU}\) in the bounding box loss function,
which utilizes measurements of intersection scale and non-maximum
suppression in post-processing to enhance object occlusion overlap
without increasing computational cost.

The goal of the \(\mathcal{L}_{CIoU}\) is to minimize this function,
enhancing the model\textquotesingle s ability to regress accurately to
target bounding boxes while accommodating variations in bounding box
shapes and sizes. Here is the mathematical representation of the
\(\mathcal{L}_{CIoU}\) and an explanation of its parameters:
\begin{ceqn}
\begin{align}
\mathcal{L}_{CIoU} = 1 - IoU + \ \frac{\rho^{2}\left( \mathbf{b},\ \ \mathbf{b}^{gt} \right)}{c^{2}} + \alpha\upsilon
\end{align}
\end{ceqn}
Where, \(IoU\) is the Intersection over Union, representing the ratio of
the intersection to the union of the two bounding boxes, commonly
referred to as IoU. \(\rho\) is the Euclidean distance between the
centers of the two bounding boxes. \(\mathbf{b}\) and
\(\mathbf{b}^{gt}\) are the center points of the two bounding boxes.
\(c\) is the length of the diagonal of the two bounding boxes.
\(\alpha\) is a hyperparameter used to balance the importance of center
point distance and aspect ratio. \(\upsilon\) is a term used to penalize
inconsistent aspect ratios.

\subsubsection{Action Recognition Network}

In the paper, we trained our all datasets by STNs based on ResNeXt
the backbone for 100 epochs and utilized the mini-batch Stochastic gradient
descent to optimize our training procedure (learning rate: $0.1$;
weight decay: 1e-5; momentum: $0.9$; batch size: 64).

The  ARN model formulation follows conventional action
methods\textsuperscript{40-42}. We set
\(X = \left\{ X_{i} \right\},i \in \lbrack 1,N\rbrack\) the mice TLE
training dataset, where \(N\) is the number of the mice videos in the
training set. Let
\(X_{i} = \left\{ x_{i1},x_{i2},\ldots,x_{iG} \right\}\) denotes
\(i_{th}\) video consisting of \(G\) non-overlapping clips with certain
frames. \(\mathcal{F}\left( x_{ij};\text{W} \right)\) represents the
function of the mice TLE prediction model with the parameters \(W\) on
an input \(x_{ij}\). The output of the model is
\(s_{ij} = \left\{ s_{ij}^{1},s_{ij}^{2},\ldots,s_{ij}^{C} \right\}\),
where \(s_{ij}^{c}\) is the prediction score of \(c_{th}\) class and
\(C\) is the number of classes. We adopt the Softmax function
\(\mathcal{S}\) to normalize the output prediction of \(\mathcal{F}\),
which can be formulated as

\begin{ceqn}
\begin{equation}
    {\bar{s}}_{ij}^{c} = \frac{e^{s_{ij}^{c}}}{\sum_{k = 1}^{C}e^{s_{ij}^{k}}}
\end{equation}
\end{ceqn}

where \({\bar{s}}_{ij}^{c}\) is the normalized score of \(s_{ij}^{c}\).

We use cross entropy (CE) loss for the loss function of the ARN model, which is formulated as

\begin{ceqn}
\begin{equation}
    \mathcal{L}(y,x,W) = - \sum_{k = 1}^{C}y_{k}\text{log}\mathcal{S}_{k}(\mathcal{F}(x;W))
\end{equation}
\end{ceqn}

where one-hot vector \(y = (y_{1},\ldots,y_{C})^{T}\) is the ground
truth label for the input \(x\), \(\mathcal{S}_{k}\) represents the
function of \(\mathcal{S}\) on \(k_{th}\) class.

To create training samples, we adhere to a widely accepted data
augmentation approach {[}1, 2{]}. Initially, we employ uniform time
sampling to randomly select a frame from a video. Subsequently, we
generate a 64-frame video clip, with the chosen frame as the starting
point. If the clip is shorter than 64 frames, we loop it as needed. In
the spatial domain, we randomly select a location from four corners and
the center, while applying a random spatial scale. The scale parameter
represents the ratio of the width of the cropped patch to the shorter
side length of the frame. The samples are randomly cropped at various
positions and scales, maintaining an aspect ratio of 1, and then resized
to dimensions of $112 \times 112$. The resulting sample is represented as $3 \times 64 \times 112 \times 112$, where $3$ denotes the
RGB channel. Following established practices, we normalize the channels
by subtracting the means and dividing by the variances of the
ActivityNet dataset. Additionally, during training, we introduce random
sample flips with a 0.5 probability. Importantly, the class labels
assigned to the generated samples remain consistent with those of their
original videos.

\subsubsection{Dataset, Training/Validation/Testing}

\begin{figure}[H]
    \centering
    \includegraphics[width=\linewidth]{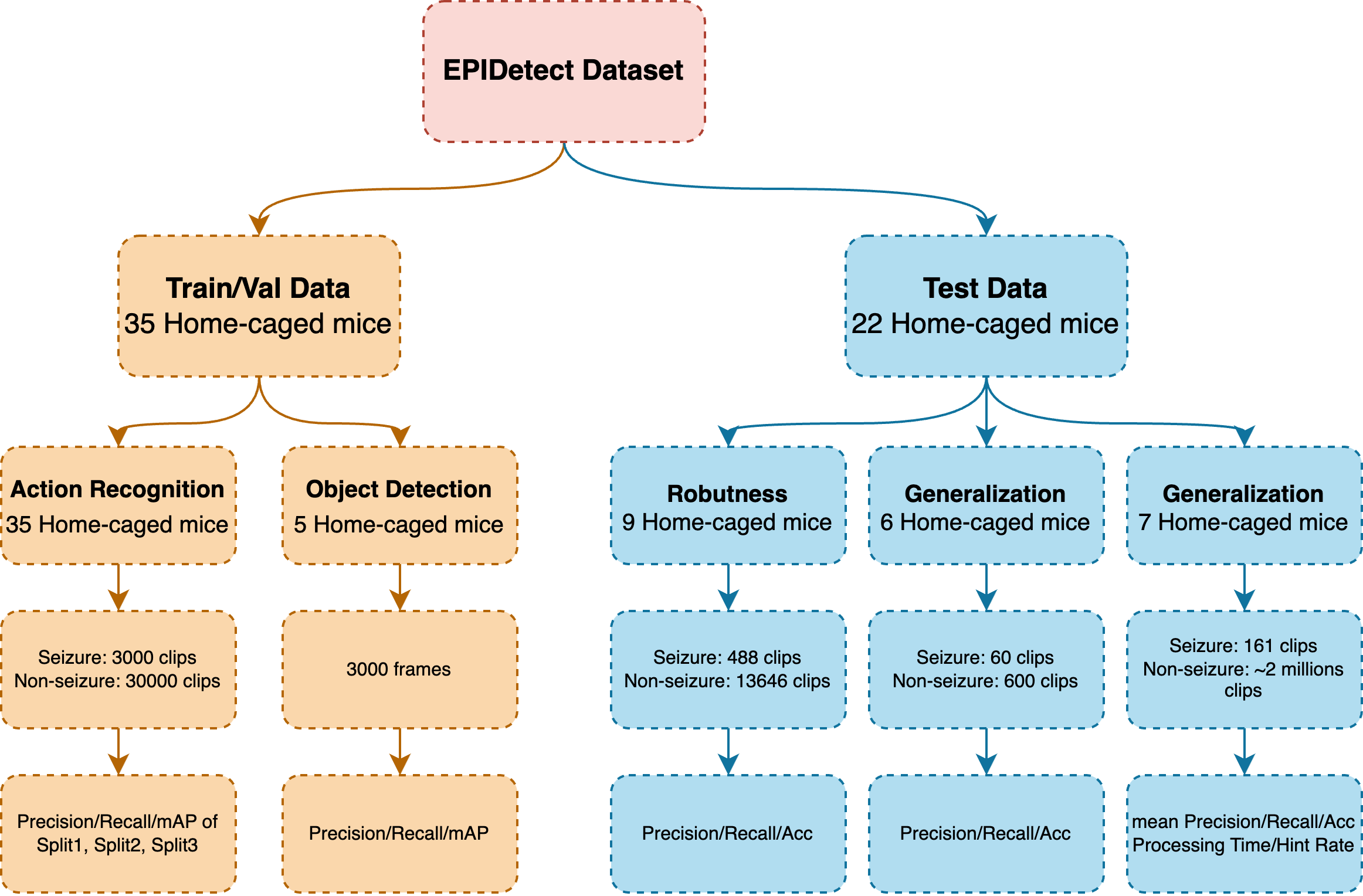}
    \caption{Training, Validation, test and evaluation in datasets for different
tasks}
    \label{fig:dataset}
\end{figure}

The EPIDetect dataset comprises two primary components: Train/Validation
and Test. Train/Validation is categorized into two distinct tasks:
object detection and action recognition. The Object Detection dataset
consists of images showcasing the mouse object of interest for
localization, while the Action Recognition dataset comprises frames
labeled as non-seizure or seizure behavior, essential for detection.
These components are integral to establishing a supervised Deep Learning
approach for training, validation, and testing, as depicted in
\textbf{Fig. \ref{fig:dataset}}.

For the Object Detection task, we extracted 3000 frames from the six 
home-caged mice to label the precise location of the mouse object as the
Region of Interest (ROI). An 80\% portion of the data was allocated for
training, with the remaining 20\% serving as the validation set.
Ensuring accurate object localization is pivotal, with the ROI defining
the animal\textquotesingle s position through a square delineated by
width, height, and coordinates (x, y). Due to the vastness of the video
recordings, which encompassed over a million frames, we annotated only a
representative subset. Frame selection was facilitated by extracting
image features using a pre-trained convolutional neural network,
specifically YOLOv5, which had been trained on the extensive MS COCO
(Microsoft Common Objects in Context) dataset \cite{lin2014microsoft}). The image annotation
process was conducted utilizing cloud services, specifically Roboflow
and Labelbox.

For Action Recognition, classification labels distinguish between
seizure and non-seizure behaviors. Seizure classification is determined
based on distinct convulsive behaviors, including myoclonic seizure
(sudden and repetitive head and neck movements, tail stiffening), clonic
seizure (forelimb clonus and rearing), and tonic-clonic seizure
(widespread running, jumping, and falling). To establish the dataset for
non-seizure clips, we initially randomly selected multiple videos from
the dataset, excluding epileptic clips. These selected clips were then
trimmed to match the duration of epileptic clips. Subsequently, we
scrutinized these trimmed clips for normal mouse behaviors, encompassing
activities such as grooming, walking, and eating.

We trained 18 individual networks, each with varying sizes of the
training set, encompassing three splits for 1\%, 5\%, 10\%, 20\%, 50\%,
and 80\% of the training set size. The best-performing models were
evaluated across different training proportions using one split, with a
20\% validation dataset for analysis and selection. To ensure the
accuracy of annotations, five experts were assigned to annotate the
Train/Validation video data, comprising 3,000 clips of seizure behavior
and 30,000 clips of non-seizure behavior, following detailed
instructions. Each expert received precise instructions for annotating
the onset and offset of epileptic behaviors, and inter-rater agreement
was assessed based on the labeling time between the two experts. This
agreement rate is computed as:
\begin{ceqn}
\begin{equation}
    R_{A_{1}A_{2}} = \frac{T_{A_{1}} \cap T_{A_2}}{T_{A_{1}} \cup T_{A_2}}
\end{equation}
\end{ceqn}
Where we defined the seizure duration that the annotator \(A_{1}\) and
\(A_{2}\) annotated in a specific video as \(T_{A_{1}}\)and
\(T_{A_{2}}\) respectively. We defined the agreement rate
\(R_{A_{1}A_{2}}\) equaled to the intersection of \(T_{A_{1}}\) and
\(T_{A_{2}}\)by the union of \(T_{A_{1}}\) and \(T_{A_{2}}\).

\begin{figure}[H]
  \begin{subfigure}{0.5\textwidth} % 0.5\textwidth will make the figures each take up half of the line width
    \centering
    \includegraphics[height=5.5cm, width=0.9\linewidth]{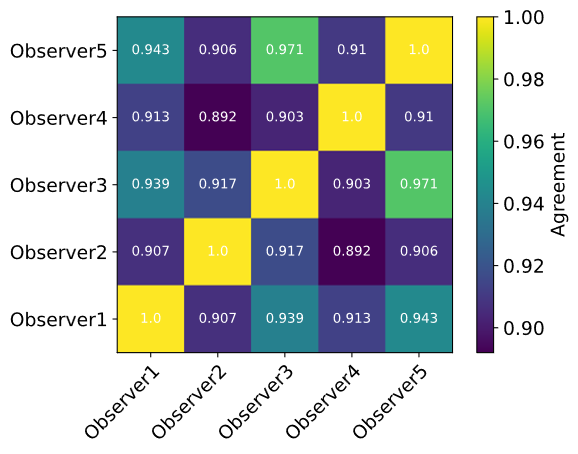}
    \caption{Agreement rate between observers}
  \end{subfigure}
  \begin{subfigure}{0.5\textwidth}
    \centering
    \includegraphics[width=0.9\linewidth]{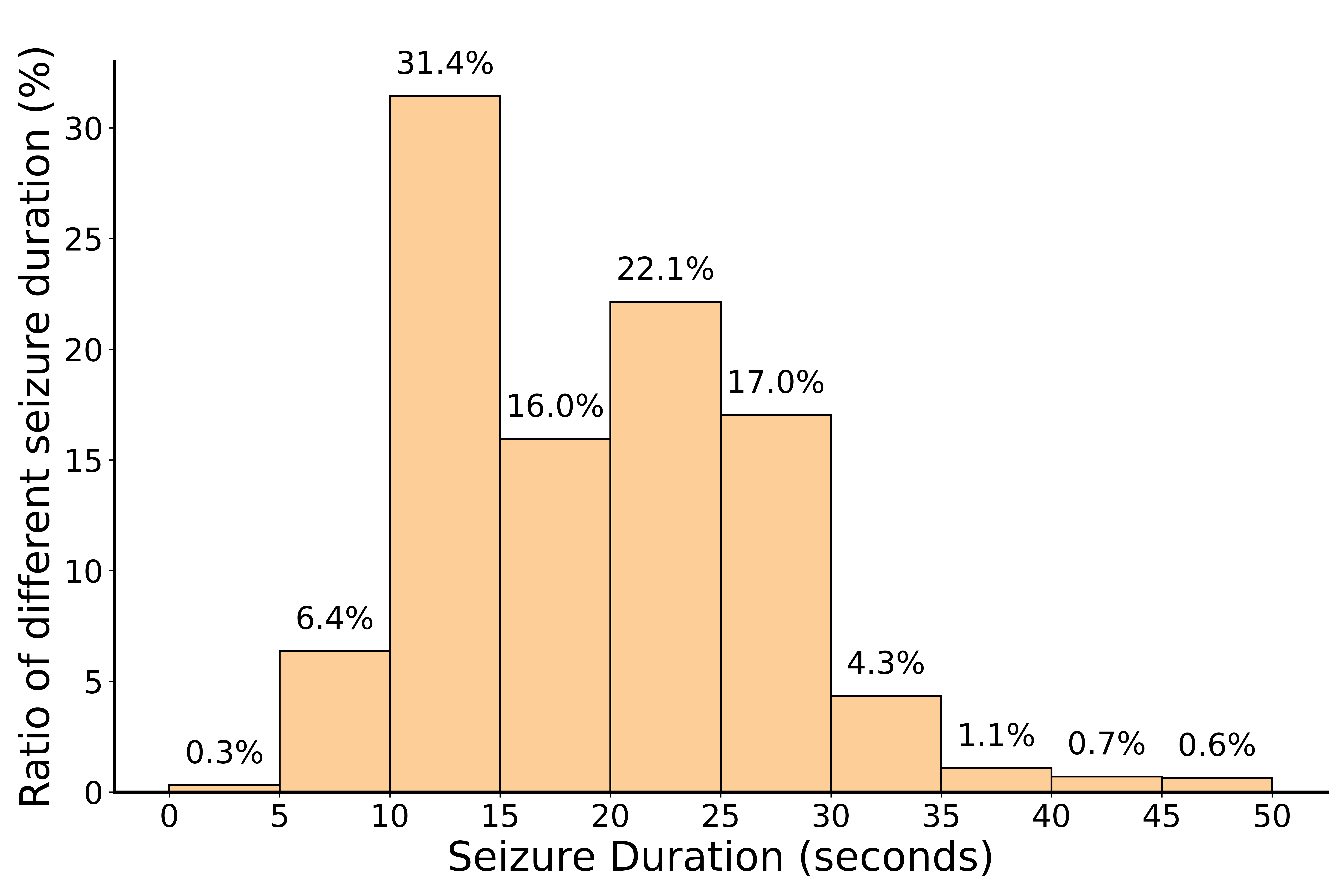}
    \caption{Convulsive seizure duration distribution}
  \end{subfigure}
  \caption{Data labeling and distribution}
  \label{fig:labelling}
\end{figure}

These experts achieved a strong average agreement of 92.0\%. the results
from the two annotators with the highest agreement rate (97.1\%) are
regarded as standard and the two annotators will annotate each video in
other datasets (\textbf{Fig. \ref{fig:labelling}} and \textbf{Fig. S1}).

% \begin{figure}[H]
%     \centering
%     \includegraphics[width=\linewidth]{Figure5.png}
%     \caption{Data labeling and distribution a) agreement rate between observers.
% b) convulsive seizure duration distribution in our dataset}
%     \label{fig:labelling}
% \end{figure}

The testing phase aims to assess the robustness, generalization, and
preclinical applicability of our method. The test data is divided into
three datasets: robustness (R-test), generalization (G-test), and preclinical drug
testing (PD-test). The R-test dataset consists of 488 separate
epileptic video clips and 13,646 non-epileptic clips, totaling 2,046,900
frames. These clips were obtained from seven mice housed in home cages
and recorded over a three-week period. Additionally, the 60
individual convulsive seizure clips and 600 non-seizure behavioral
clips from six epileptic mice in experimental chambers are annotated as G-test dataset to evaluate the
generalizability of our methods. The evaluation of results includes
metrics such as precision, recall, and accuracy. To compare EPIDetect
with human experts in a real-world setting, a PD-test dataset
comprising 148 continuous videos recorded over one whole day from seven
epileptic mice is used to test the anti-epileptic efficacy of a cholecystokinin B receptor antagonist-YF476.

\subsubsection{Evaluation of object detection and action recognition}

For the object detection task, only one class is labelled because the object
detection is for preprocessing frames in videos and mouse localization
in the home cage. Therefore, the following metrics are calculated for
correct bounding boxes and the number of correct detections:

\begin{equation}
\text{Precision} = \frac{\text{True Positives (TP)}}{\text{True Positives (TP)} + \text{False Positives (FP)}} 
\end{equation}
\begin{equation}
    \text{Recall} = \frac{\text{True Positives (TP)}}{\text{True Positives (TP)} + \text{False Negatives (FN)}}
\end{equation}

This refers to the model correctly detecting an object and accurately
locating it. If the model\textquotesingle s bounding box has sufficient
overlap with the actual object\textquotesingle s bounding box (e.g., IoU
greater than a threshold), then that box can be considered a True
Positive. This refers to the model incorrectly detecting an object or
generating bounding boxes in areas without objects. These boxes either
do not overlap with the actual object\textquotesingle s bounding box or
have insufficient overlap. False Negative (FN): This refers to the model
failing to detect an object that is actually present. If the actual
object\textquotesingle s bounding box is not detected by the model,
it\textquotesingle s considered a False Negative. You can use these
metrics to evaluate the performance of your single-class object
detection model. Usually, an IoU threshold is chosen to define what kind
of bounding box is considered correct. A higher IoU threshold (e.g., 0.5
or higher) requires stricter bounding box matching, while a lower IoU
threshold (e.g., 0.3) allows some degree of overlapping bounding boxes
to be considered correct. Mean Average Precision (mAP) is represented as
the mean value of the Average Precision (AP) calculated for different
classes or categories. The formula to compute mAP involves summing up
the AP values for each class and then dividing by the total number of
classes.

Mathematically, for $N$ classes:
\begin{ceqn}
\begin{equation}
    \text{mAP} = \frac{1}{N} \sum_{i=1}^{N} AP_i 
\end{equation}
\end{ceqn}
Here, $AP_i$ represents the Average Precision for class $i$, and $N$ is the total
number of classes. In essence, mAP provides an overall measure of a
model\textquotesingle s detection performance across different classes,
balancing the precision-recall trade-off for each class.

For action recognition, the precision and recall metrics are the same as
object detection task but the mAP is defined as the mean value of the Average
precision from three different split datasets. The area under curve (AUC)
of the precision-recall curve as AP to indicate whether the models could
correctly detect all positive samples without incorrectly marking so
many non-seizures as seizure.

The F1-Score is the harmonic mean of Precision and Recall, used to
balance both metrics. It can be calculated using the formula:
\begin{ceqn}
\begin{equation}
    F1 = \ \frac{2 \cdot Precision \cdot Recall}{Precision + Recall}
\end{equation}
\end{ceqn}

The test dataset is evaluated similarly to action recognition while the
preclinical test belongs to real-environment without labeling and
calculated by processing time and hint rate. The hint rate is calculated as:
\begin{ceqn}
\begin{equation}
    \text{Hint\ rate} = \ \frac{N_{1}}{N_{1} \cup N_{2}}\ or\ \frac{N_{2}}{N_{1} \cup N_{2}}
\end{equation}
\end{ceqn}

Where $N_{1}$ is the detected seizure by EPIDetect, \(N_{2}\ \)is the
detected seizure by human experts. \(N_{1} \cup N_{2}\) is the total
detected seizure by EPIDetect and human experts.

% \section{experimental set up}
\subsection{Implementation of the methodology in splitting train/val datasets}

\subsubsection{YOLOv5 localizes epileptic mice in home cages}

In the initial phase, 3000 frames were labelled, with 2400 allocated for
training and 600 for validation. These labels encompassed the mouse
behavior annotation and the corresponding ROI.

The Pytorch architecture is leveraged for the implementation of YOLOv5 for
300 epochs of training and validation. Pertained weights of YOLOv5s from
the MS COCO dataset are incorporated for network operation, running on a
per-frame basis to predict the mouse\textquotesingle s position.

\textbf{Fig. S2} offers a visual depiction of the training procedure for
object localization. Although the mouse becomes a smaller object due to
moving to a deeper depth in the home cage, YOLOv5s could learn the correct ROI and
localize the mouse position.

\begin{figure}[H]
    \centering
    \includegraphics[width=\linewidth]{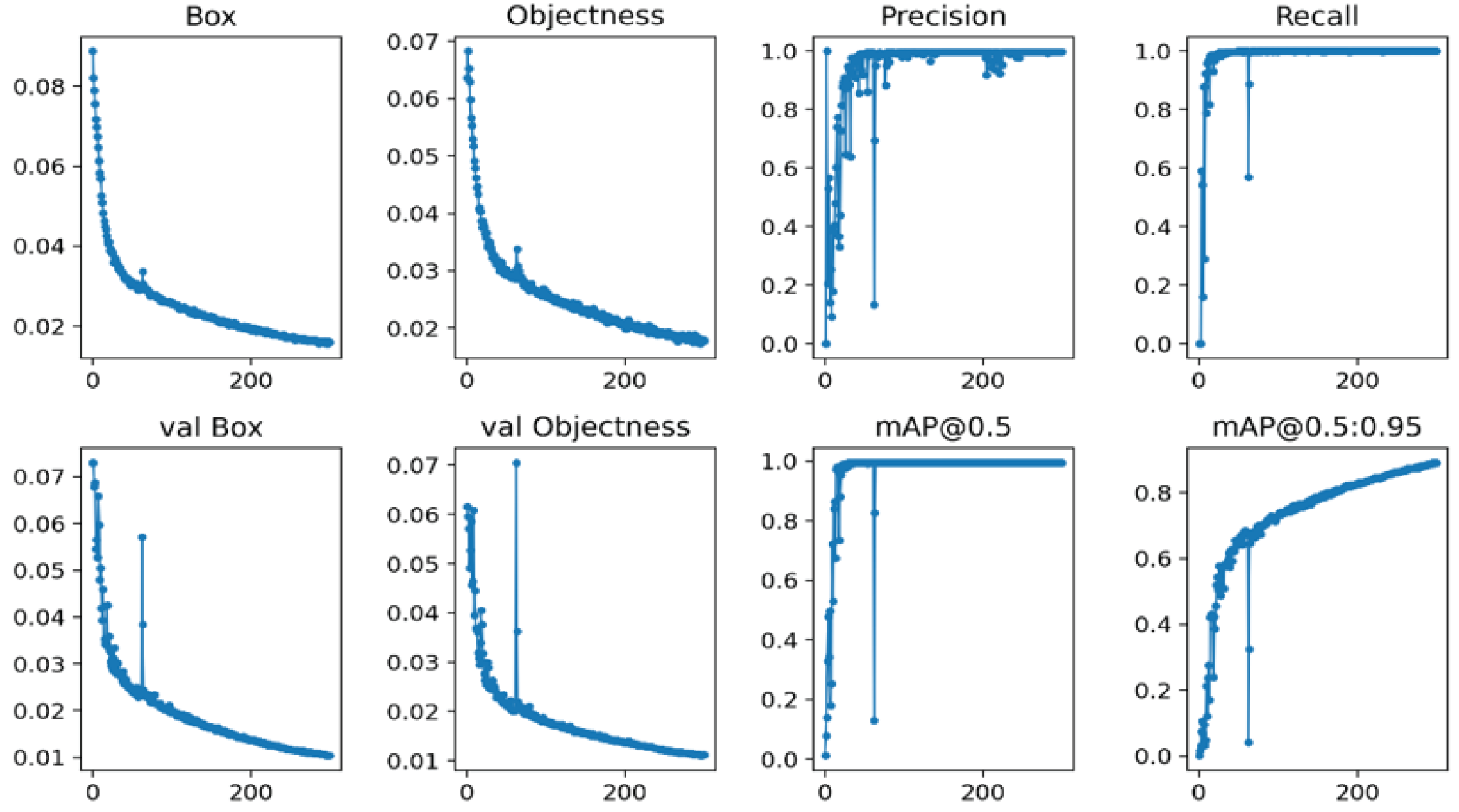}
    \caption{YOLOv5 performance in our home-caged dataset}
    \label{fig:YOLOv5}
\end{figure}

Within the validation dataset, precision and recall scores of 0.99 and
0.99, respectively, are attained (\textbf{Fig. \ref{fig:YOLOv5}}). In the validation
dataset, the network attains mAP of 0.99 under the intersection over the Union
(IoU) of 0.5 and mAP of 0.85 across the IoU threshold range from 0.5 to
0.95, signifying its accuracy in correctly localizing the animal within
the image. However, these results fall short of our objectives,
particularly considering the balanced nature of the validation dataset,
which is inconsistent with practical application scenarios.
Consequently, this network is exclusively utilized for ROI prediction.

\subsubsection{3d-ResNext detects the convulsive seizures in home-caged mice}

\begin{figure*}[t]
  \centering
  \begin{subfigure}{0.4\textwidth}
    \centering
    \includegraphics[height=5.5cm, width=\linewidth]{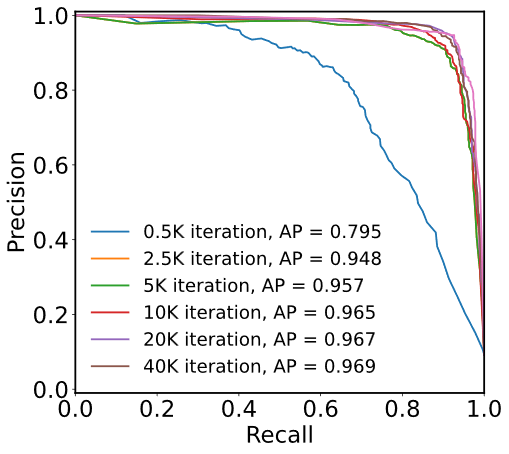}
    \caption{PR Curve of different Iteration}
  \end{subfigure}
  \begin{subfigure}{0.4\textwidth}
    \centering
    \includegraphics[height=5.5cm, width=\linewidth]{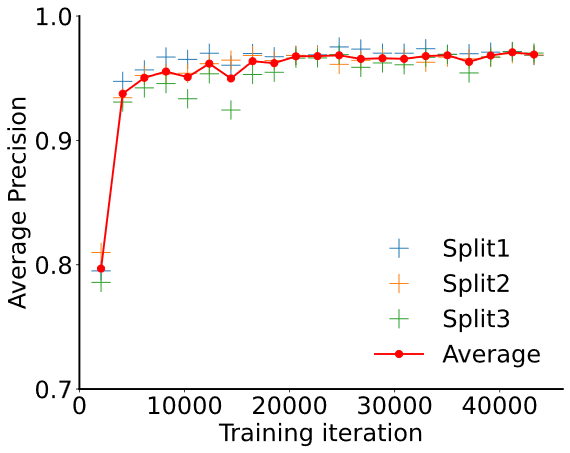}
    \caption{AP of different Iteration}
  \end{subfigure}\\
  \begin{subfigure}{0.4\textwidth}
    \centering
    \includegraphics[height=5.5cm, width=\linewidth]{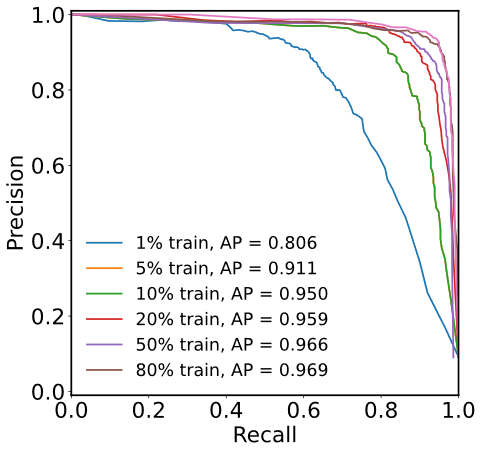}
    \caption{PR Curve of different Ratio}
  \end{subfigure}
  \begin{subfigure}{0.4\textwidth}
    \centering
    \includegraphics[height=5.5cm, width=\linewidth]{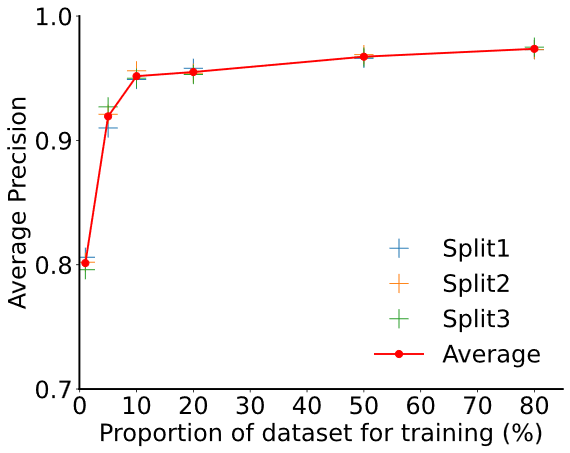}
    \caption{AP of different ratio}
  \end{subfigure}
  \caption{The performance of models on different training/validation sets is represented using precision-recall (PR) curves and average precision (AP) scores}
  \label{fig:PRPlot}
\end{figure*}

We conducted a random split of the dataset into three segments: training
(80\%), validation (20\%), and tested the ARN\textquotesingle s
performance over various training iterations. Additionally, we trained
networks with different depths to compare their performance on
validation datasets (\textbf{Table. \ref{tab:ar}}). The mean Average Precision results for three identical splits of the different proportional
training set fractions (ResNeXt-18: 66.7\%, ResNeXt-50: 68.7\%,
ResNeXt-101: 78.2\%), with the 101-layer network outperforming the
others, achieving a 78.2\% mAP at a frame rate of 16. Furthermore,
models based on networks with a frame rate of 64 frames surpassed those
with a frame rate of 16 frames, with the 101-layer models achieving an
85.0\% mAP. A transfer learning technique is applied to initialize the
weights of our neural network instead of training from scratch. We used
a large pre-trained model from the extensive human action dataset,
Kinetics-400, which includes over 65 million frames and 400 human action
labels, to fine-tune our dataset with the same neural network. The
fine-tuned models at different depths outperformed both networks trained
from scratch, with the 101-layer model achieving a remarkable 96.8\%
mAP. This demonstrates our ability to achieve exceptional performance in convulsive seizure detection in home-caged mice by leveraging transfer learning from Kinetics-400 (K400). We also conducted an analysis of the correlation
between performance, training iterations, and the proportion of the
dataset for 3d-ResNext-101 at a 64-frame rate. The precision-recall
curve illustrates the average precision change based on one split of
validation data for different iterations, ranging from 500 to 40,000
(\textbf{Fig. \ref{fig:PRPlot}}). The performance for the three different set splits
improved, with mAP increasing from 79.68\% to 96.8\% (\textbf{Fig. \ref{fig:PRPlot}}).
Similarly, the performance in the validation dataset exhibited
significant improvements as the training dataset proportion increased
from 1\% to 80\%.

\begin{table}[H]
    \centering
    \caption{Performance of AR in train/val 80\%/20\%}
    \begin{tabular*}{\linewidth}{@{\extracolsep{\fill}}cccc}
        \toprule
        \textbf{Method} & \textbf{Backbone} & \textbf{Frames} & \textbf{Val mAP} \\
        \midrule
        \multirow{6}{*}{ Scratch } & \multirow{2}{*}{18} & 16 & 65 \\
        & & 64 & \textbf{66.7} \\
        \cmidrule{2-4}
        & \multirow{2}{*}{50} & 16 & 68.7 \\
        & & 64 & \textbf{72.5} \\
        \cmidrule{2-4}
        & \multirow{2}{*}{101} & 16 & 78.2 \\
        & & 64 & \textbf{85.0} \\
        \midrule
        \multirow{3}{*}{Kinetics-400} & 18 & 64 & 83.6 \\
        & 50 & 64 & 93.4 \\
        & 101 & 64 & \textbf{96.8} \\
        \bottomrule
    \end{tabular*}
    \label{tab:ar}
\end{table}

\begin{figure*}[t]
  \begin{subfigure}{0.5\textwidth}
    \centering
    \includegraphics[height=5.5cm, width = \linewidth]{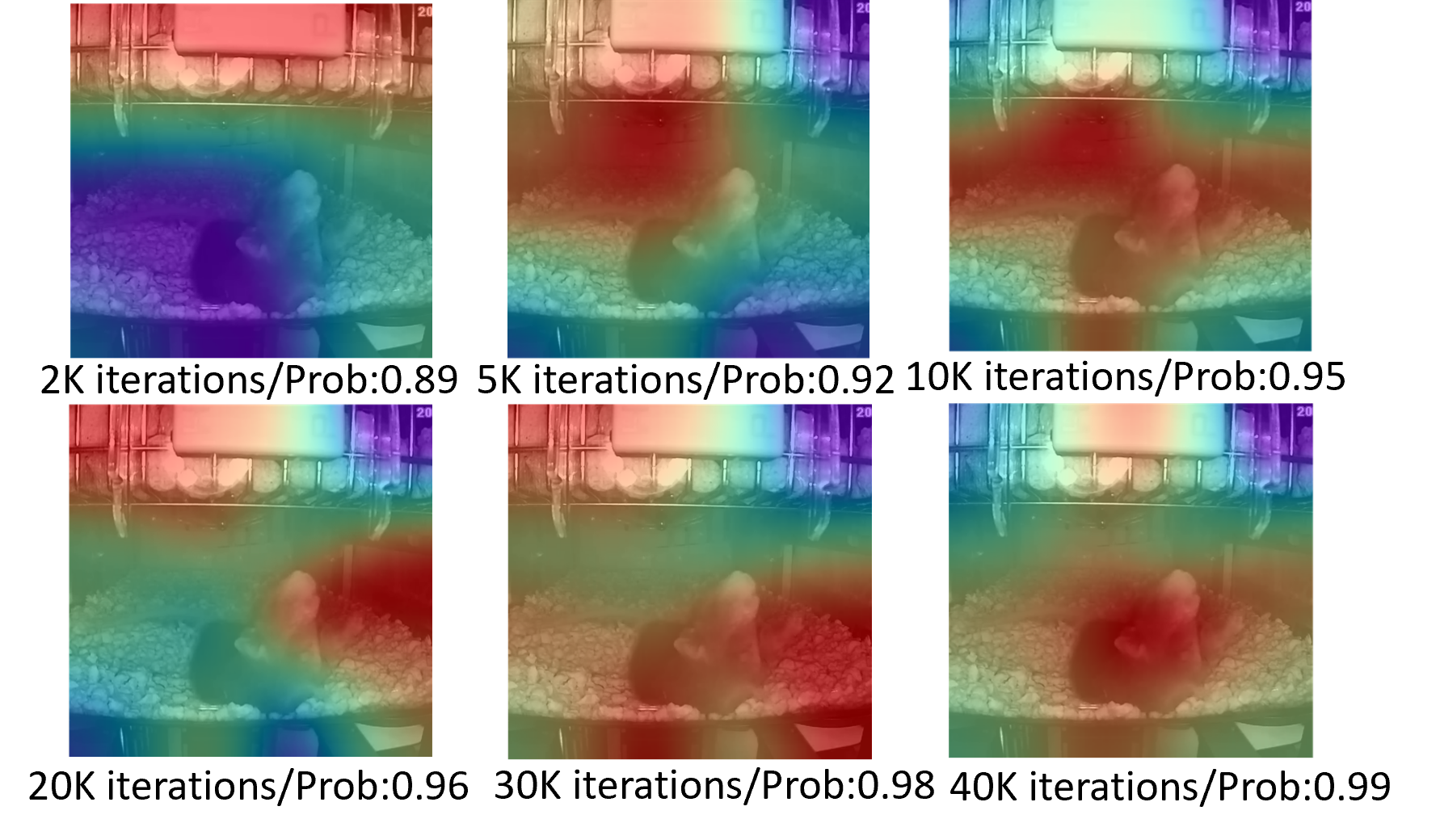}
    \caption{Different training iterations}
  \end{subfigure}
  \begin{subfigure}{0.5\textwidth}
    \centering
    \includegraphics[height=5.5cm, width=\linewidth]{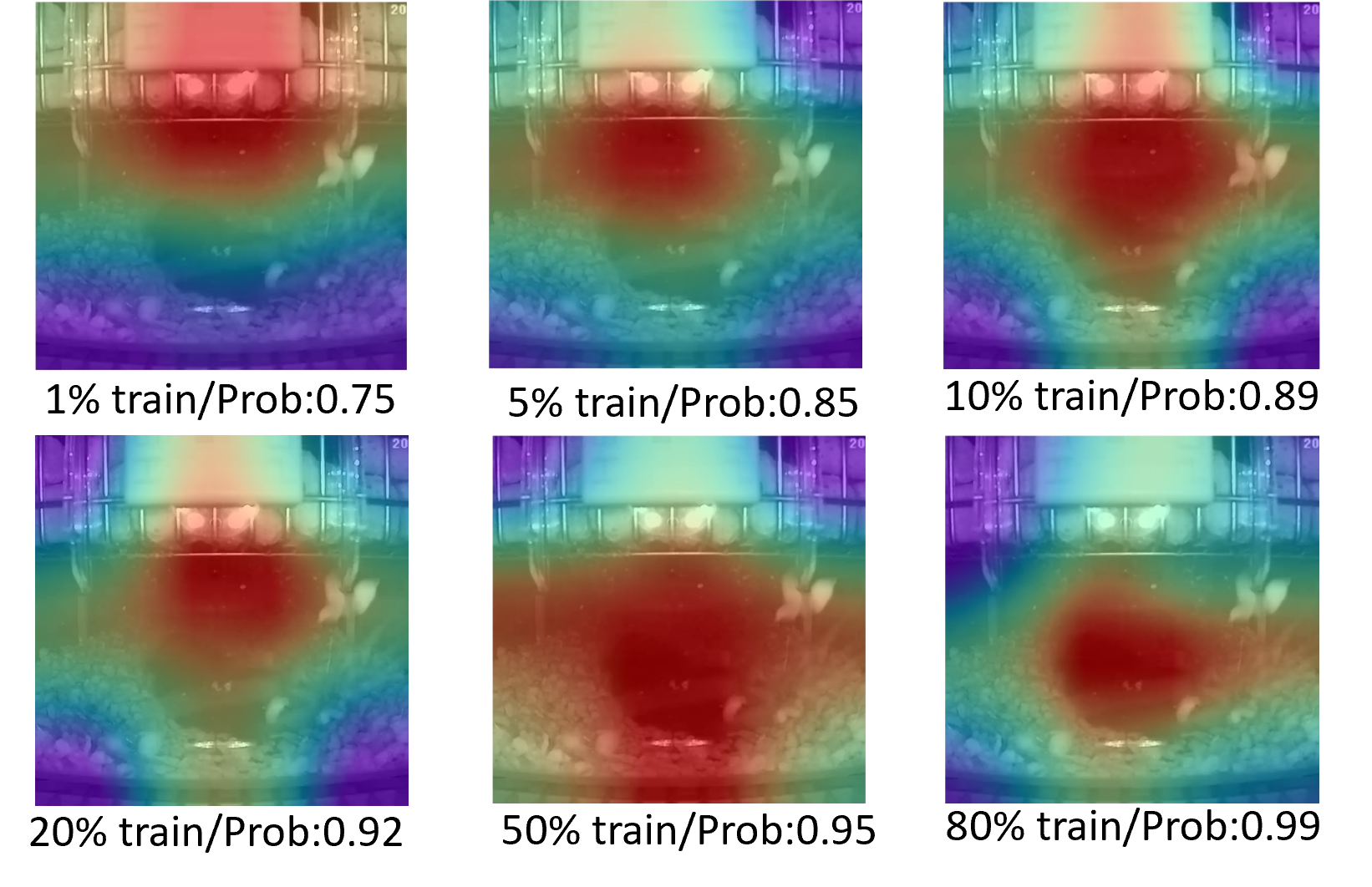}
    \caption{Different training proportion}
  \end{subfigure}
  \caption{Example of visualized Grad-CAM in one seizure video with different training iterations and training proportions. a) the change of Grad-CAM from 2000 to 40000 training iterations. b) the change of Grad-CAM from $1\%$ to $80\%$ training proportion.}
  \label{fig:CAM}
\end{figure*}

As a tool for explaining the decision-making process of deep-learning networks, Grad-CAM visualizes the areas of focus within the network on video frames to analyze object and action classification. In this context, Grad-CAM was employed to illustrate how the network directs its attention to home-caged mice in a video, varying with increasing training iterations and training proportions (\textbf{Fig. \ref{fig:CAM}}). Within this video, the prediction probability for seizures increased from 0.89 to 0.99 as the iterations progressed from 2000 to 40000. This demonstrates that the model became well-trained on our dataset during the training process. Similarly, across different proportions of training data, Grad-CAM also reveals an increase in prediction probability from 0.75 to 0.99 as the proportion of training data increases from $1\%$ to $80\%$. This indicates that the network is learning more features of seizure behaviors as the number of video data increases.

% \begin{figure}[H]
%     \centering
%     \includegraphics[width=2.59097in,height=3.15139in]{Figure8.png}
%     \caption{Example of visualized Grad-CAM in one seizure video with different training iterations and training proportions. a) the change of Grad-CAM from 2000 to 40000 training iterations. b) the change of Grad-CAM from $1\%$ to $80\%$ training proportion.}
%     \label{fig:CAM}
% \end{figure}

% \begin{figure}[H]
%   \begin{subfigure}{0.5\textwidth} % 0.5\textwidth will make the figures each take up half of the line width
%     \centering
%     \includegraphics[height=5.5cm, width=\linewidth]{Figure8a.png}
%     \caption{Different training iterations}
%   \end{subfigure}
%   \begin{subfigure}{0.5\textwidth}
%     \centering
%     \includegraphics[height=5.5cm, width=\linewidth]{Figure8b.png}
%     \caption{Different training proportion}
%   \end{subfigure}
%   \caption{Example of visualized Grad-CAM in one seizure video with different training iterations and training proportions. a) the change of Grad-CAM from 2000 to 40000 training iterations. b) the change of Grad-CAM from $1\%$ to $80\%$ training proportion.}
%   \label{fig:CAM}
% \end{figure}

% The author names and affiliations could be formatted in two ways:
% \begin{enumerate}[(1)]
% \item Group the authors per affiliation.
% \item Use footnotes to indicate the affiliations.
% \end{enumerate}
% See the front matter of this document for examples. 
% You are recommended to conform your choice to the journal you 
% are submitting to.

\section{Results}

\subsection{The performance of centre crop and object detection crop}

\begin{figure}[t]
    \centering
    \includegraphics[width=\linewidth]{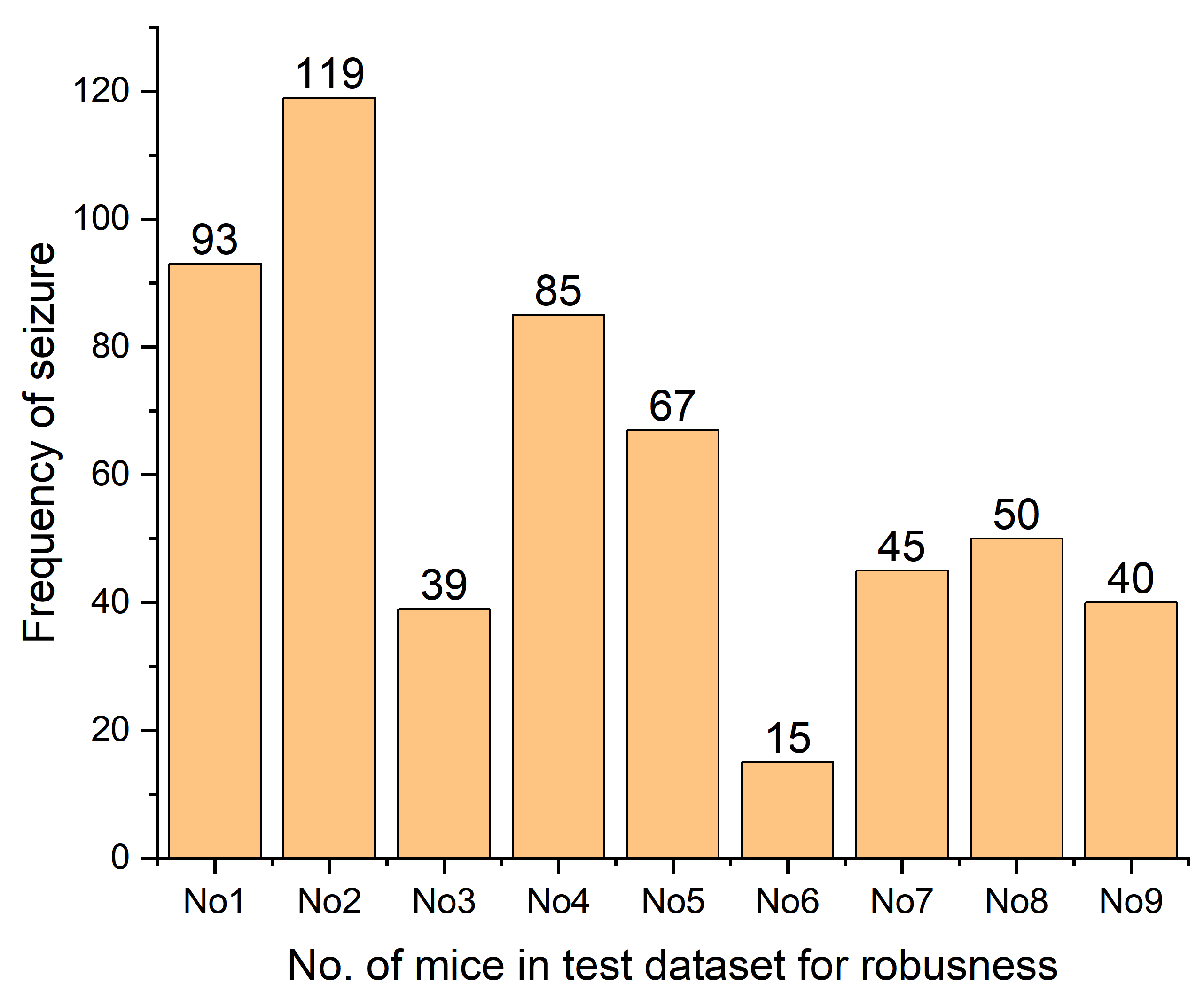}
    \caption{The number of seizures and non-seizures in different mice}
    \label{fig:seizure_freq}
\end{figure}

The ARN of the EPIDetect platform demonstrates its capability to detect convulsive seizures in the validation dataset; however, its ability to generalize to unseen objects remains uncertain. Additionally, it's unclear whether there's any performance improvement when replacing centre cropping with ODN and other strategies. To address these challenges, we first evaluate the model's performance on a test dataset to assess its robustness. This R-test dataset consists of trimmed videos, including 553 separate epileptic video clips and 15,318 non-epileptic clips (with a total of 2,046,900 frames) obtained from 9 mice kept in home cages and recorded for three weeks (\textbf{Fig. \ref{fig:seizure_freq}}). These mice exhibited varying seizure frequencies during the observation period, ranging from 15 to 119 seizures. The ARN models trained on three different splits of the training dataset using 3d-ResNext-101 are employed to detect seizures and non-seizures in the R-test dataset.

\begin{figure}[H]
    \centering
    \includegraphics[width=\linewidth]{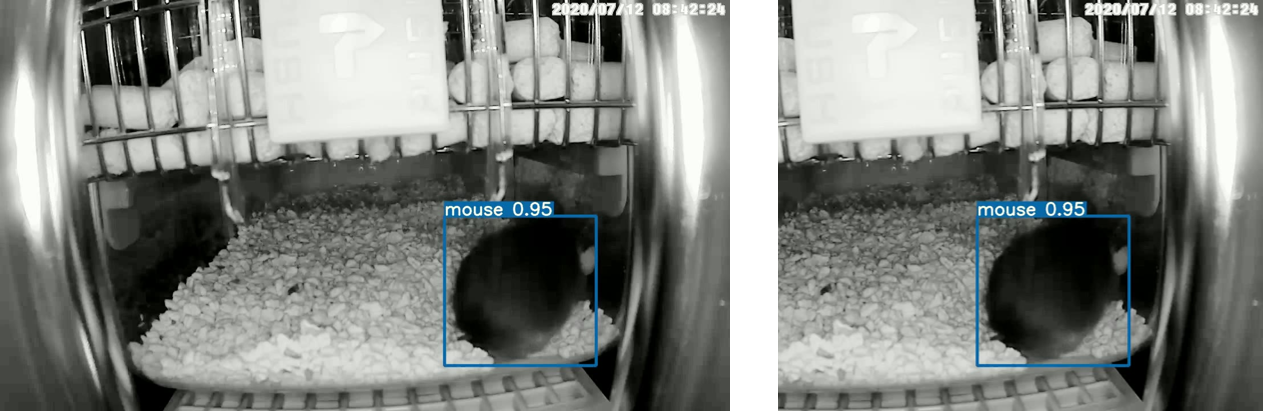}
    \caption{Illustration of object detection crop (OD crop) }
    \label{fig:center_crop}
\end{figure}

\begin{figure}[H]
    \centering
    \includegraphics[width=\linewidth]{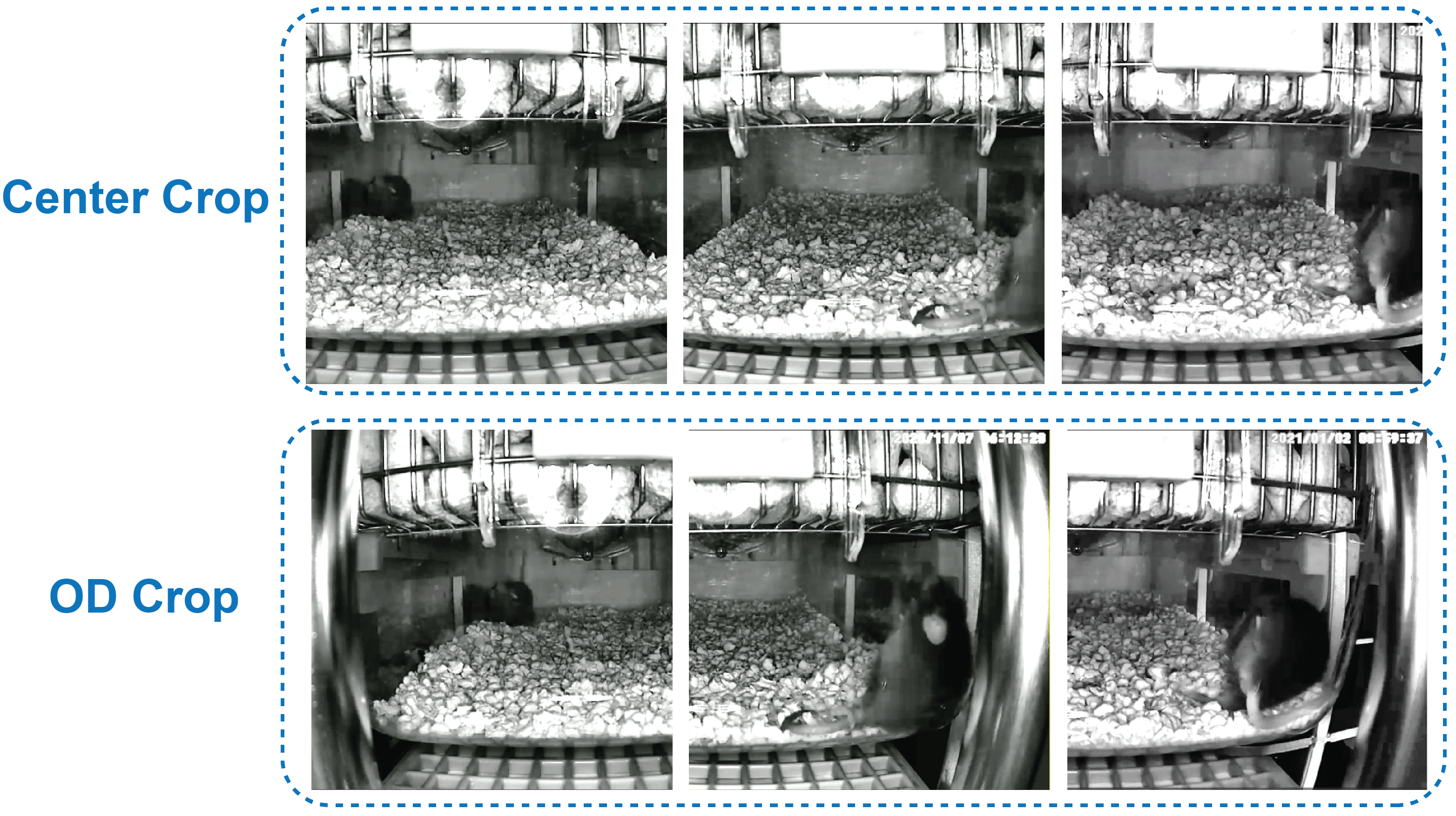}
    \caption{The cropped image difference with centre crop or OD crop during the inference phase}
    \label{fig:different_crop}
\end{figure}

The centre crop has been commonly used as a preprocessing method for video input during the inference phase. However, this approach has limitations as it may result in incomplete object detection, leading to unstable detection performance when the object moves to the edge of the video frame due to varying proportions of length and width. This issue can be addressed by incorporating object detection as a preprocessing step. Instead of centre cropping to obtain a square input based on the shorter side, we propose cropping the square input using the centre coordinates of the target object box, guided by the ODN. This approach ensures that the mouse remains in the centre of the input frame, making it easier to detect (\textbf{Fig. \ref{fig:center_crop}}). The performance of ARN with ResNext-101 backbone and object detection crop (OD crop) is compared with this ARN with a YOLOv5s backbone and centre crop. In three video examples, it's evident that the localization of mice using OD crop in ARN outperforms centre cropping in the ARN (\textbf{Fig. \ref{fig:different_crop}}).

\subsection{The Performance comparison of seizure detection in R-test dataset in different decision-making strategies}

Furthermore, the results of the R-test dataset demonstrate notable improvements in ARN performance when utilizing OD crop in comparison to centre crop (\textbf{Table. \ref{tab:test_dataset})}. Specifically, the results achieved with the OD crop yielded an impressive F1 score of 98.7\% in split1, surpassing the 98.5\% achieved with the centre crop. In pursuit of further enhancing the EPIDetect system, we introduced parallel prediction using models trained on three different split datasets. Additionally, we explored two post-processing methods: simple average and weighted average. In the context of the simple average, we computed the sum of predicted probabilities based on models from the three split training datasets. Weighted Average, on the other hand, involved multiplying the predicted probabilities of the three models by their respective weights and then normalizing them by the sum of weights, where each weight corresponds to the model's performance on the validation dataset.
Remarkably, both the simple average and weighted average approaches yielded similar results when employed in ARN with OD crop (Recall: 98.9\%; Precision: 99.1\%; F1 score: 99.0\%), and it is noteworthy that OD crop consistently outperformed centre crop in both of these post-processing methods.

\begin{table*}[t]
    \centering
    \begin{threeparttable}
    \caption{Model performance of different split training data in R-test dataset}
    \begin{tabularx}{0.7\linewidth}{cXccc}
    \toprule
    Model & Method & Recall(\%) & Precision(\%) & F1 score(\%) \\
    \midrule
    \multirow{2}{*}{Split1} & Centre crop &  98.9 & 98.2 & 98.5 \\
    & OD crop & 98.9 & 98.6 & \textbf{98.7} \\
    \midrule
    \multirow{2}{*}{Split2} & Centre crop &  96.6 & 98.8 & 97.7 \\
    & OD crop & 96.6 & 99.1 & \textbf{97.8} \\
    \midrule
    \multirow{2}{*}{Split3} & Centre crop &  95.3 & 98.9 & 97.1 \\
    & OD crop & 96.1 & 98.9 & \textbf{97.5} \\
    \midrule
    \multirow{2}{*}{Simple Average} & Centre crop & 98.7 & 99.1 & 98.9 \\
    & OD crop & 98.9 & 99.1 & \textbf{99.0} \\
    \midrule
    \multirow{2}{*}{Weight Average} & Centre crop & 98.8 & 99.0 & 98.9 \\
    & OD crop & 98.9 & 99.1 & \textbf{99.0} \\
    \bottomrule
    \end{tabularx}
    \begin{tablenotes}
    \small
    \item Note: The values in the table are in percentages. ``BOLD''
represents a better value compared with another one.
    \end{tablenotes}
    \label{tab:test_dataset}
    \end{threeparttable}
\end{table*}

In this context, we further assessed the performance of the YOLO3d101-WA approach in the EPIDetect platform, which combines ARN with 3d-ResNext-101, OD crop using YOLOv5s, and a weighted average of results from three splits, on the R-test dataset while varying the threshold parameters. We specifically examined probabilities predicted as 'seizure' at thresholds of 0.2, 0.5, and 0.8 to classify seizure and non-seizure clips across nine different home-caged mice in the R-test dataset.
\textbf{Table. \ref{tab:YOLO_test}} provides a summary of the results, indicating that as the threshold increases, the recall rate either decreases or remains constant at 100\%, while the precision rate increases or remains consistent. Notably, a threshold value of 0.5 yielded the most balanced trade-off between recall and precision, resulting in the highest average F1 score of 99.0. Consequently, users have the flexibility to adjust the threshold value according to their specific requirements to attain different objectives.
However, it's worth noting that home-caged mice No.6 and No.8 exhibited particularly low precision rates of 60.0\% and 49.0\%, respectively, at a threshold of 0.2. This drop in precision may be attributed to factors such as background illumination in the video, recording direction, and the limited availability of positive samples.
\begin{table*}[t]
    \centering
    \caption{YOLO3d101 model performance in different mice of R-test dataset with threshold adjustment}
    % \begin{tabularx}{\textwidth}{l*{9}{X}}
     % \begin{tabularx}{\textwidth}{>{\raggedright\arraybackslash}p{1.8cm}*{9}{>{\centering\arraybackslash}X}}
    \begin{tabularx}{\textwidth}{>{\raggedright\arraybackslash}p{1.8cm}*{9}{>{\centering\arraybackslash}m{1.25cm}}}
    \toprule
     & \multicolumn{3}{c}{Recall(\%)} & \multicolumn{3}{c}{Precision(\%)} & \multicolumn{3}{c}{F1 score(\%)} \\
    % \midrule
    Mouse & 0.2 & 0.5 & 0.8 & 0.2 & 0.5 & 0.8 & 0.2 & 0.5 & 0.8 \\
    \midrule
     No.1  & $\mathbf{1 0 0}$ & $\mathbf{1 0 0}$ & $\mathbf{1 0 0}$ & 96.9 & 97.9 & $\mathbf{1 0 0}$ & 98.4 & 98.9 & $\mathbf{1 0 0}$ \\
     No.2 & $\mathbf{1 0 0}$ & $\mathbf{1 0 0}$ & $\mathbf{1 0 0}$ & $\mathbf{9 9 . 2}$ & 98.3 & 95.0 & $\mathbf{9 9 . 6}$ & 99.2 & 97.4 \\
     No.3 & $\mathbf{1 0 0}$ & $\mathbf{1 0 0}$ & 97.4 & 92.9 & 95.1 & $\mathbf{1 0 0}$ & 96.3 & 97.5 & $\mathbf{9 8 . 7}$ \\
     No.4 & $\mathbf{9 8 . 8}$ & $\mathbf{9 8 . 8}$ & $\mathbf{9 8 . 8}$ & 96.6 & $\mathbf{1 0 0}$ & $\mathbf{1 0 0}$ & 97.7 & $\mathbf{9 9 . 4}$ & $\mathbf{9 9 . 4}$ \\
      No.5 & $\mathbf{1 0 0}$ & 98.5 & 97.0 & 94.3 & $\mathbf{1 0 0}$ & $\mathbf{1 0 0}$ & 97.0 & $\mathbf{9 9 . 2}$ & 98.5 \\
     No.6 & $\mathbf{1 0 0}$ & $\mathbf{1 0 0}$ & $\mathbf{1 0 0}$ & 60.0 & $\mathbf{1 0 0}$ & $\mathbf{1 0 0}$ & 75.0 & $\mathbf{1 0 0}$ & $\mathbf{1 0 0}$ \\
     No.7 & $\mathbf{1 0 0}$ & $\mathbf{1 0 0}$ & 95.6 & 97.8 & $\mathbf{1 0 0}$ & $\mathbf{1 0 0}$ & 98.9 & $\mathbf{1 0 0}$ & 97.7 \\
     No.8 & $\mathbf{1 0 0}$ & 98.0 & 94.0 & 49.0 & 98.0 & $\mathbf{1 0 0}$ & 65.8 & $\mathbf{9 8 . 0}$ & 96.9 \\
     No.9 & $\mathbf{1 0 0}$ & 97.5 & 97.5 & 90.9 & $\mathbf{1 0 0}$ & $\mathbf{1 0 0}$ & 95.2 & $\mathbf{9 8 . 7}$ & $\mathbf{9 8 . 7}$ \\
    Average & $\mathbf{9 9 . 9}$ & 99.2 & 97.8 & 86.4 & 98.8 & $\mathbf{9 9 . 4}$ & 91.5 & $\mathbf{99 . 0}$ & 98.6 \\
    \bottomrule
    \end{tabularx}
    \label{tab:YOLO_test}
\end{table*}

``Avg'' presents the Average data value of nice home-caged mice.

\subsection{Prediction and visualization in EPIDetect system}

\begin{figure}[H]
    \centering
    \includegraphics[width=\linewidth]{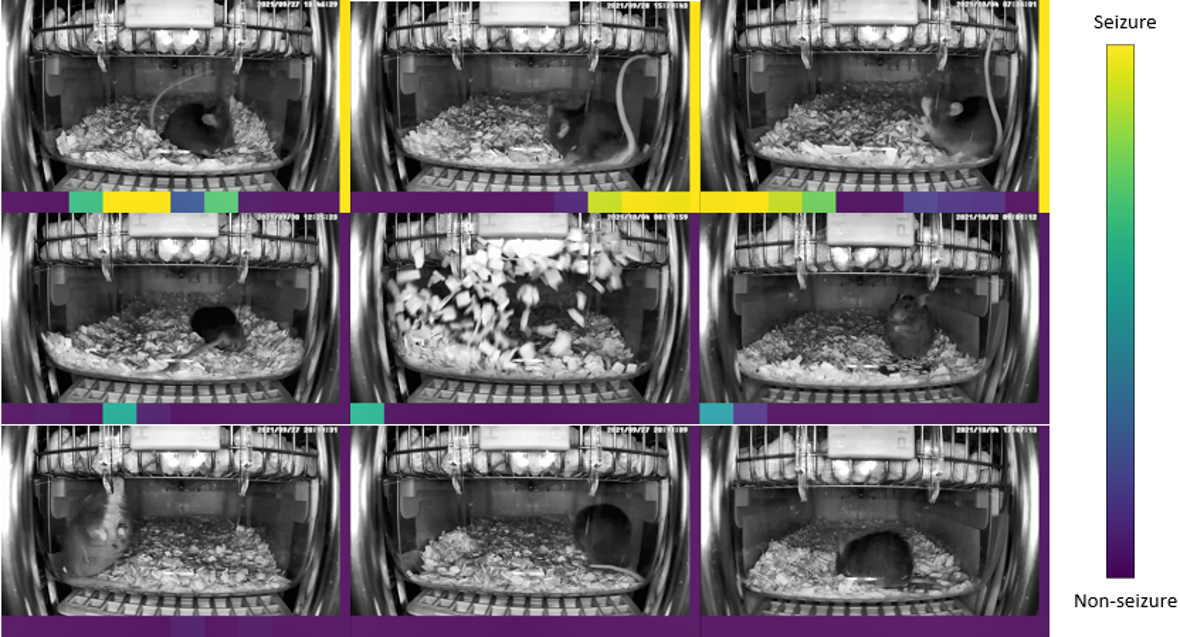}
    \caption{Visualization example of nine one-minute videos including seizure samples and non-seizure samples}
    \label{fig:enter-label}
\end{figure}

We provide a visual representation of decision threshold adjustments using a 10X prediction approach, where one-minute videos are predicted, resulting in ten viewpoints, each comprising 64 frames (depicted at the bottom). EPIDetect's YOLO3d101-WA model generates predictions based on these viewpoints, distinguishing between convulsive seizures and spontaneous seizure behaviors in mice. The colour bar in the right column illustrates whether the entire video clip is classified as a seizure (yellow) or non-seizure (purple). This analysis includes nine one-minute videos: three containing seizure video clips (first row) and six containing non-seizure video clips (second and third rows) (\textbf{Fig. \ref{fig:enter-label}} \textbf{, Supplementary videos}).

At a threshold of 0.5, seizure video clips are correctly identified as seizures, while three non-seizure clips in the second row are incorrectly classified as seizures, and those in the third row are correctly identified as non-seizures. However, there can be instances where one or more viewpoints yield incorrect predictions. By increasing the threshold to 0.8, we observe a noticeable reduction in false positive results while maintaining high accuracy in detecting epileptic and non-epileptic behaviors. This demonstrates that raising the threshold to 0.8 can enhance performance, particularly by reducing the number of false positives in the classification of epileptic and non-epileptic behaviors.

\subsection{Drug efficacy test using EPIDetect system}

To further assess the practical utility of EPIDetect in preclinical drug efficacy testing, the PD-test dataset comprises unlabelled continuous video clips spanning three weeks from seven epileptic mice housed in home cages. This dataset included a seven-day pre-treatment period, followed by seven days of treatment with a newly discovered anti-epileptic drug called YF476, and finally, a seven-day post-treatment period. Our team of annotators was tasked with annotating epileptic behaviors and counting the frequency of seizures from these videos.~

\begin{figure*}[t]
    \centering
    \includegraphics[width=\linewidth]{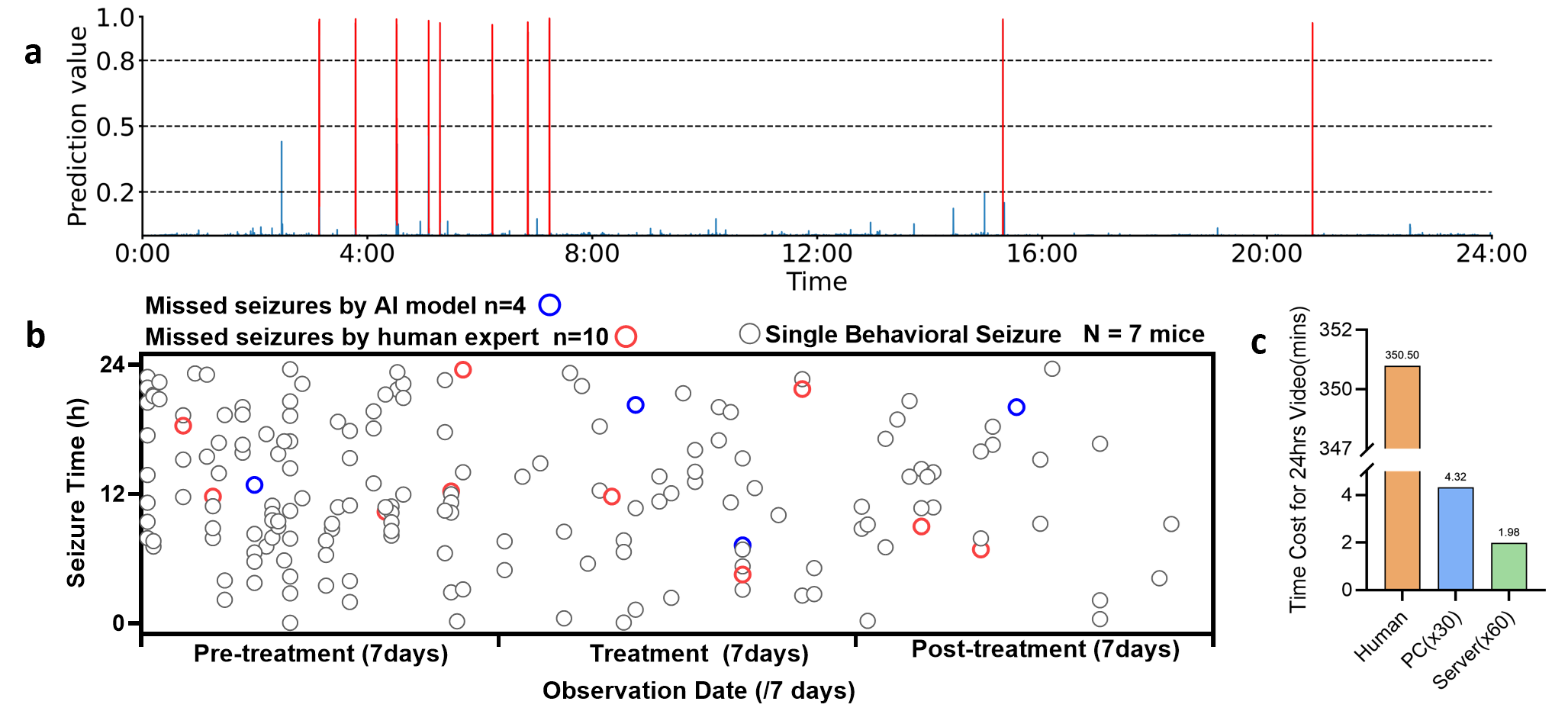}
    \caption{EPIDetect platform based on YOLO3d101 test real-world data to compare with expert. a) Practical application performance in 24 hours. b) Practical application performance in large-scale drug screening. c) Time consumption comparison between EPIDetect and human expert.}
    \label{fig:compare_expert}
\end{figure*}

To evaluate the efficacy of EPIDetect in practical applications, we employed the EPIDetect system with YOLO3d101-WA and a threshold of 0.5 (YOLO3d101-WA-0.5) to analyze this dataset. The results for a one-day segment showed the prediction probability for seizures, with the red line indicating seizures and the blue line indicating non-seizures. Notably, clear distinctions between seizures and non-seizures were observed \textbf{(Fig. \ref{fig:compare_expert}a)}. Additionally, EPIDetect processed a 24-hour video in just 1.98 minutes on our server, representing a 175-fold increase in processing speed compared to human expert observation \textbf{(Fig. \ref{fig:compare_expert}b)}. Preliminary findings indicated a reduction in the frequency of seizures in epileptic mice following our treatment regimen (convulsive seizure frequency: pre-treatment 90; treatment 43; post-treatment 28) \textbf{(Fig. \ref{fig:compare_expert}c)}. Furthermore, our automated approach detected more epileptic seizures (n=4) than human experts (n=10). The seizure detection accuracy of the automated approach (97.5\%) surpassed that of human observation (93.8\%) (seizure detection accuracy is calculated as the convulsive seizure count detected by our approach or human experts divided by the total seizure count detected by our approach and human experts). In summary, these results underscore that EPIDetect not only offers superior efficiency but also higher accuracy compared to human observers in real-time seizure detection, making it a valuable tool in preclinical drug efficacy assessment.~

\subsection{Generalization of the different environment in EPIDetect system}

\begin{figure}[H]
    \centering
    \includegraphics[width=\linewidth]{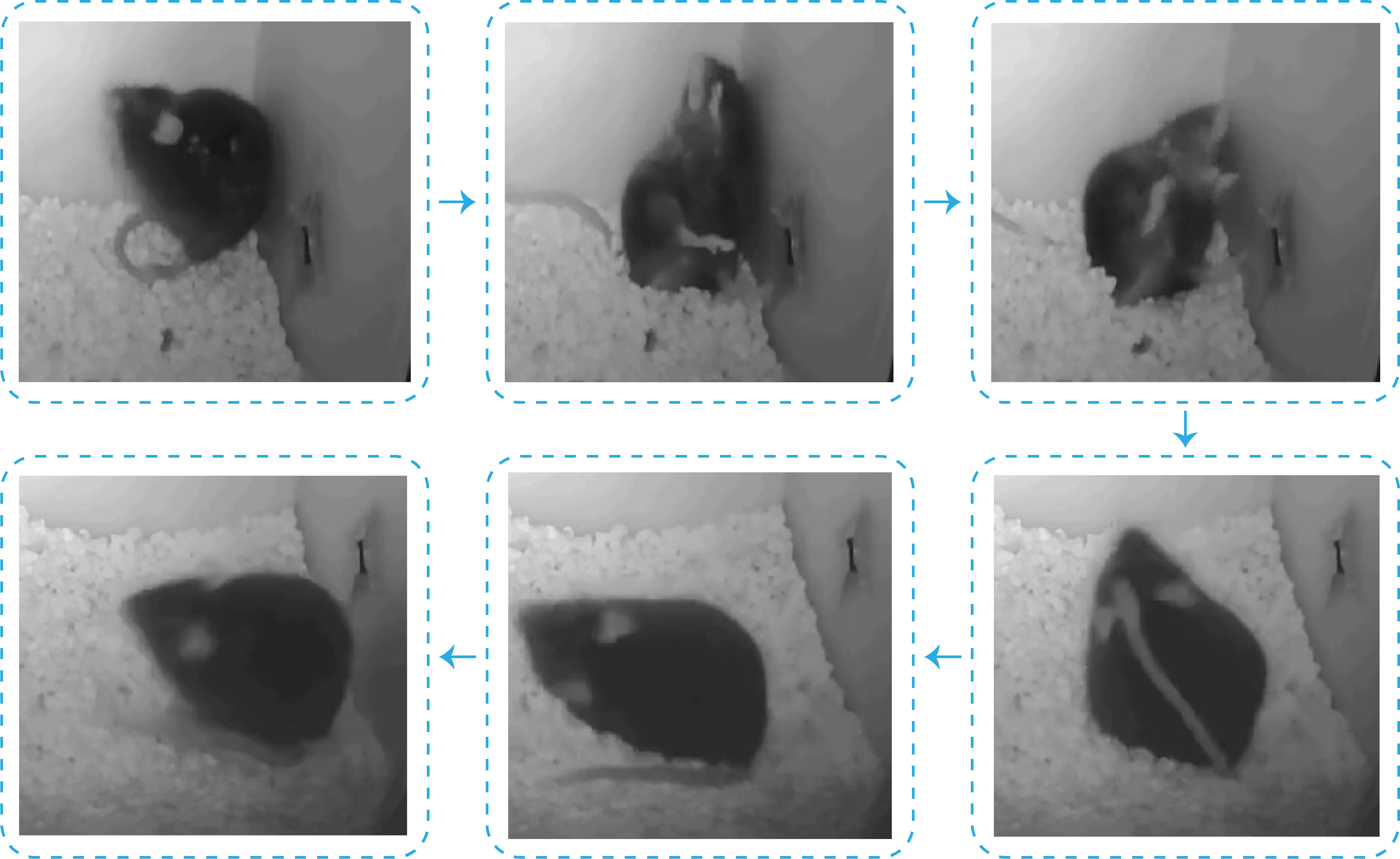}
    \caption{Convulsive seizure behavior in experimental chamber mice}
    \label{fig:seizure_behavior}
\end{figure}

The G-test dataset includes Six IHKA mice that were housed in experimental chambers for three days, during which we recorded 60 individual seizure and 600 non-seizure behavioral clips to assess the generalizability of EPIDetect ~\textbf{(Fig. \ref{fig:seizure_behavior})}. From this dataset, it is evident that convulsive seizure behaviors are rarely affected by different environmental conditions. The YOLO3d101-WA-0.5 model was employed to directly evaluate the generalization dataset (G-test) and achieved outstanding performance (Precision: 91.3\%; Recall: 95.5\%; F1 score: 93.3\%) ~\textbf{Table. \ref{tab:without_pretrain}}). Moreover, it is noteworthy that human experts spent approximately 70.2 hours labelling the data, whereas EPIDetect completed the task in just 0.55 hours. This highlights the superior efficiency of EPIDetect in processing large-scale convulsive seizure videos.

\begin{table}[h]
    \caption{This is a test caption. This is a test caption. This is a test caption. This is a test caption.}
    % \label{tbl1}
    \begin{tabular*}{\linewidth}{@{\extracolsep{\fill}}cccccc}
        \toprule
        \multicolumn{1}{c}{Method} &
        \multicolumn{1}{c}{Threshold} &
        \multicolumn{1}{c}{Precision} &
        \multicolumn{1}{c}{Recall} &
        \multicolumn{1}{c}{F1 score} \\
        \midrule
        YOLO3d101-WA & 0.5 & 91.3 & 95.5 & 93.3 \\
        \bottomrule
    \end{tabular*}
    \label{tab:without_pretrain}
\end{table}

\section{Discussion}

Abnormal behaviors resulting from neurological diseases, particularly epilepsy, exhibit significant complexity and dynamics, in contrast to the less complex spontaneous behaviors such as resting, eating, and drinking. Consequently, the time and resources required for manual data collection of epileptic behaviors pose a challenge to large-scale data analysis. Our method offers scalability, speed, and the capability to extract detailed information from data. The innovative combination of data collection methods (home-cage recording and neuroethology-based expert annotation) with seizure detection techniques (automated behavioral classification and transfer learning) can revolutionize industrial environments.
The first large-scale dataset, which includes IHKA mice, holds immense potential for advancing mechanism research and drug development, ultimately enhancing human living standards. Moreover, as the pioneering automated method for seizure detection in home-caged environments, EPIDetect simplifies the acquisition of precise data and reduces the need for extensive human involvement in behavioral observation. This can redirect funding towards other critical areas, including investigational new drug (IND) development.
However, EPIDetect does have areas for improvement. The need to periodically move the home cage from the recording area for environmental maintenance, such as providing fresh bedding, food, and water, could potentially impact detection performance, as our models do not identify individual mice for behavior classification. Another limitation lies in the dataset's imbalance between normal and non-epileptic activities. For example, grooming behaviors involve mice licking various parts of their bodies. These behaviors are non-frequent and may occasionally lead to the misclassification of normal behaviors as epileptic seizures by the machine. Additionally, detecting seizures with lower intensity (below three degrees) remains challenging due to their high similarity to everyday activities. This challenge could potentially be addressed by incorporating EEG data for annotating synchronized video clips in future iterations of the method.
Furthermore, each step in our pipeline is not specific to the epileptic seizure behaviors of animal models, making our method adaptable to other animals and behaviors associated with neurodegenerative diseases. We envision future applications of our pipeline in diverse contexts, such as behavioral analysis in primates within three-dimensional spaces and the study of conditions like depression, Parkinson's disease, and Alzheimer's disease.

\section{Conclusion~and future work}

We showcase the versatility of our deep learning framework, EPIDetect, in the context of preclinical anti-epilepsy treatment evaluation. Our framework excels across multiple dimensions, including performance (comparable to human expertise), scalability (handling dozens to thousands of epileptic videos), experimental subjects (mice), and environments (home cages and chambers). EPIDetect utilizes models pre-trained with various architectures and offers comprehensive guidelines for:
\begin{enumerate}
\def\labelenumi{\arabic{enumi}.}
\item
  Automate video trimming based on labels of user-define,
\item
  Data preprocessing, including enhancement and augmentation,
\item
  Predicting seizures in new home-caged mice or chambered mice,
\item
  Inference for unlabeled videos, facilitating the screening of new anti-epilepsy drugs.
\end{enumerate}

Furthermore, the models trained on the Kinetics-400 dataset, as presented in this paper, can aid users in fine-tuning or predicting results for their specific experimental data. Our models exhibit the capability to generalize across unseen epileptic mice recorded from different cameras. Additionally, we demonstrate the adaptability of our approach by showcasing how performance varies with different threshold values for epileptic prediction.
In real-time applications, our methods significantly reduce the processing time for 24 hours of video data, reducing it from 350.5 minutes to a mere 1.98 minutes (1/175 of the original duration). This remarkable speed enhancement outperforms human experts in terms of epileptic detection robustness. Furthermore, we emphasize the potential for detecting abnormal behaviors, even those of lower seizure degrees, in future iterations of our framework.

\section{Acknowledgement}

This work was supported by the ITC: Drug Discovery in Regenerative Medicine (2020-2025) and ITF (GHP/075/19GD): Platform development for new anti-epileptic drug evaluation and pre-clinical study of new cholecystokinin antagonists as new anti-epileptic drugs (01/06/2021-31/05/2023), which both are funded by Innovation and technology commission of the government of Hong Kong. 

\bibliographystyle{unsrtnat}
% \bibliographystyle{cas-model2-names}
% \bibliographystyle{model1-num-names}
% Loading bibliography database
%\bibliography{cas-refs}

\end{document}